\documentclass[11pt]{article}

\usepackage[preprint]{acl}

\usepackage{times}
\usepackage{latexsym}
\usepackage[T1]{fontenc}
\usepackage[utf8]{inputenc}

\usepackage{microtype}

\usepackage{inconsolata}

\usepackage{amsmath}

\usepackage{todonotes}
\usepackage{amssymb}
\usepackage{multirow}
\usepackage{listings} % Include the listings package
\usepackage{xcolor} % Optional, for customizing colors

\newtheorem{definition}{Definition}

\lstdefinestyle{mystyle}{
    commentstyle=\color{codegreen},
    keywordstyle=\color{magenta},
    numberstyle=\tiny\color{codegray},
    stringstyle=\color{codepurple},
    basicstyle=\ttfamily\footnotesize,
    breakatwhitespace=false,         
    breaklines=true,                 
    captionpos=b,                    
    keepspaces=true,                 
    numbersep=5pt,                  
    showspaces=false,                
    showstringspaces=false,
    showtabs=false,                  
    tabsize=2
}

\newcommand{\SUMMARY}[1]{\todo[color=lightgray,inline]{#1}}

\newcommand{\CODERL}[0]{\text{CodeRL{}}}
\newcommand{\COPILOT}{\text{Copilot{}}}
\newcommand{\GITHUBCOPILOT}[0]{\text{GitHub \COPILOT{}}}
\newcommand{\GPT}[0]{\text{GPT-3.5}}
\newcommand{\GPTFOUR}[0]{\text{GPT-4.0}}

\newcommand{\CLLaMA}[0]{\text{Code Llama}}
\newcommand{\Mistral}[0]{\text{Mistral}}

\title{Impeding LLM-assisted Cheating in Introductory Programming Assignments via Adversarial Perturbation}

\author{Saiful Islam Salim\thanks{Authors contributed equally}, Rubin Yuchan Yang\footnotemark[1], Alexander Cooper\footnotemark[1], \\ {\bf Suryashree Ray}, {\bf Saumya Debray}, {\bf Sazzadur Rahaman\thanks{Corresponding author} } \\
        University of Arizona, Tucson, AZ, USA \\ 
        \textit{\{saifulislam, yuchan0401, alexanderecooper, suryashreeray, debray, sazz\}@arizona.edu}
}

\begin{document}
\maketitle

\begin{abstract}
While Large language model (LLM)-based programming assistants such as CoPilot and ChatGPT can help improve the productivity of professional software developers, they can also facilitate cheating in introductory computer programming courses. Assuming instructors have limited control over the industrial-strength models, this paper investigates the baseline performance of 5 widely used LLMs on a collection of introductory programming problems, examines adversarial perturbations to degrade their performance, and describes the results of a user study aimed at understanding the efficacy of such perturbations in hindering actual code generation for introductory programming assignments. The user study suggests that \textit{i)} perturbations combinedly reduced the average correctness score by 77\%, \textit{ii)} the drop in correctness caused by these perturbations was affected based on their detectability.
\end{abstract}

\section{Introduction}\label{sec:intro}

%%% Sazz:
%%% Cite this paper: https://arxiv.org/pdf/2402.11702.pdf

Large Language Model (LLM)-based tools such as ChatGPT~\cite{chatGPT} have demonstrated an impressive ability to create high-quality code given simple prompts and have the potential for significant impact on software development \cite{DBLP:journals/pacmpl/BarkeJP23}.  While there are ongoing efforts to incorporate such tools into computer science (CS) education \cite{Jacques23}, integrating new technologies into educational curricula 
can take a long time \cite{hembree1986effects, onlineEdu2022}. Meanwhile, existing CS curricula are under the threat of LLM-assisted cheating and require immediate attention \cite{DBLP:conf/ace/Finnie-AnsleyDL23, DBLP:conf/ace/Finnie-AnsleyDB22}.

\begin{figure}[ht]
    \centering
    \includegraphics[width=\linewidth]{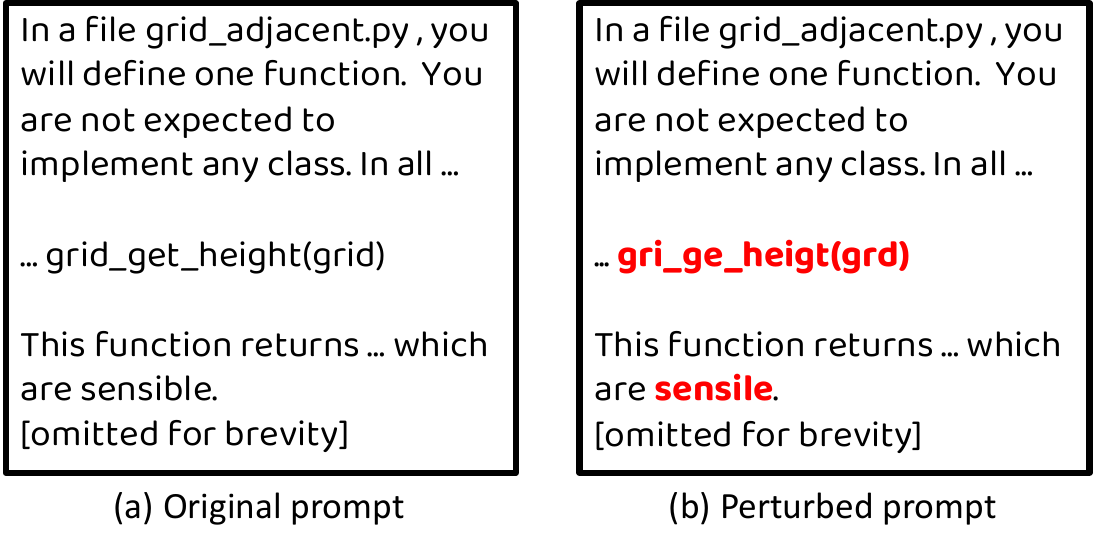}
    % \vspace{-20pt}
    \caption{Removal of 5 characters from an assignment prompt caused correctness scores of the generated solutions to drop from 100\% to 0\% in \CODERL, \CLLaMA, \GPT, and \GITHUBCOPILOT. For \Mistral{}, it dropped from 33.33\% to 0\%.}
    \label{fig:intro_motivation_figure}
\end{figure}

Given that educators have little direct control over the capabilities of industrial-strength LLMs, two possible directions towards addressing this threat are $(i)$ to detect and penalize LLM-assisted cheating; and $(ii)$ to modify problem statements to impede LLM-assisted cheating. The first approach is problematic because it can be difficult to determine reliably whether some given content is LLM-generated or not \cite{DBLP:conf/aied/HoqSLBLA23, DBLP:journals/corr/abs-2307-07411}, and both false positives and false negatives are possible. In this paper, we explore the second option and ask the following question: \textit{How can instructors modify assignment prompts to make them less amenable to LLM-based solutions without impacting their understandability to students?}

% The scope of the problem and the efficacy of possible mitigations---i.e., what kinds of introductory programming problems are easy to solve using LLM-based tools, how instructors may modify their problem statements to make them less amenable to LLM-based solutions, and what kinds of problem modifications are easy for students to work around---are largely unexplored. This paper aims to bridge this gap by investigating the potential and challenges of adversarial strategies for impeding LLM-assisted cheating.

While there has been some work on the impact of adversarial prompts on LLMs~\cite{wang2023on,chen2023LLMAttack}, 
we are not aware of any research investigating
%% the problem of investigating the potential and challenges of 
adversarial strategies for impeding LLM-assisted cheating in a Blackbox setting in an academic context. To systematically study the problem, we break it into the following three steps:

\begin{description}
    \item [Step 1.] Measure the accuracy of LLMs on introductory CS programming assignments, as introductory assignments are at imminent risk~\cite{DBLP:conf/ace/Finnie-AnsleyDL23}.

    \item [Step 2.] Develop adversarial techniques to perturb programming assignment prompts and analyze their impact on the quality of LLM-generated solutions to those problems.

    \item [Step 3.] Run a user study
    %%-based field experiment 
    to understand the potential for such perturbation techniques in impeding \textit{actual} LLM-assisted cheating, focusing in particular on whether students can detect and reverse such perturbations.
    
\end{description}

\begin{figure*}
    \centering
    \includegraphics[width=\textwidth]{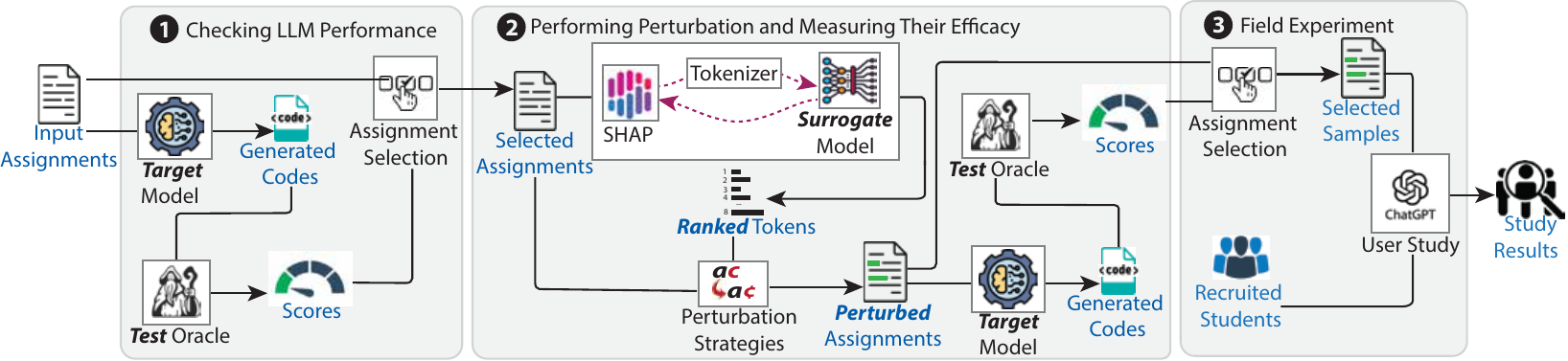}
    \caption{Overview of our study, which is conducted in three steps. Here, boxed elements indicate \fbox{processing units}, and unboxed elements represent input/output data. We used solid \textit{arrows} through processing units to connect inputs to their corresponding outputs.}
    \label{fig:overview:framework}
    \vspace{-10pt}
\end{figure*}
An overview of these steps is presented in Figure~\ref{fig:overview:framework}. To measure the accuracy of LLM-generated code, we use the same test inputs used to evaluate student submissions. To modify problem statements in a Blackbox setting, we design a set of perturbation techniques that are informed by existing literature on adversarial perturbation~\cite{DBLP:conf/icml/BielikV20,DBLP:journals/corr/RauberBB17,DBLP:conf/nips/WangXWG0GA021,DBLP:journals/corr/abs-2305-16934}. We use SHAP~\cite{SHAP_LundbergL17} with a \textit{surrogate} model to guide the perturbation for better~\textit{efficacy} vs. modification tradeoff. We define \textit{efficacy} (Definition \ref{efficacy}) for a perturbation technique to quantify the portion of lowering the LLM accuracy. To ethically conduct the user study in Step 3, we select the study group from students who have already taken the courses corresponding to the assignments used for the study.

Our findings suggest that existing LLMs generally struggle to solve assignments requiring interactions across multiple functions and classes. Our evaluation of different perturbation techniques shows a high overall success rate, causing degradation of more than 85\% of the assignments for all five models (example in Figure~\ref{fig:intro_motivation_figure}). We find that high variations in solution generations strongly correlate with high success rates. 
Our user study with undergraduates shows that the average efficacy dropped from 15.43\% to 15\% when perturbations were noticed. It also suggests that subtle perturbations, i.e., substituting tokens or removing/replacing characters, when unnoticed, are likely to retain high efficacy in impeding actual solution generation. Additionally, the \textit{detectability} of a \textit{high-change} perturbation might not imply \textit{reversion}. 
The implication is that under perturbations, students have to check and modify LLM solutions rather than adopt them unchanged -- instructors can use these perturbations when preparing homework problems to reduce cases where students do not learn but use ChatGPT as is.

\section{Measuring LLM Performance (Step 1)}\label{sec:methodology}
The goal of this evaluation is to answer the following question: \textit{How do LLMs perform on our corpus of programming assignment problems? What problems are more amenable to LLM-assisted cheating?}
% \SR{This Section goes beyond 1.5 page budget.}

\subsection{Methodology}

\noindent
\textbf{Dataset Selection and Preparation.}
For this study, we select programming assignments from the first two CS courses (CS1 and CS2) at the University of Arizona. %\sout{a large public university.}\footnote{Institution and course names are elided for reviewing.}
These courses offer problem-solving-oriented Python programming assignments focusing on basic control structures, data structures, and algorithms (Appendix \ref{appen:syllabus_cs1} and \ref{appen:syllabus_cs2}).
The assignments were designed by the instructors from the ground up, although we acknowledge that variants of the assignments may exist elsewhere, and previous students of the courses could have uploaded the assignments to the internet.
%  CS1 is designed for students with little or no prior programming experience, and CS2 is intended for students with some programming experience (not necessarily in Python).
% The programming assignments reflect this difference: the CS1 assignments are typically simpler and more specific in their requirements than CS2, e.g., with respect to the data and/or control structures students are required to use.
%% They explore the broader applicability of relevant concepts in Computer Science. Topics include arrays, lists, stacks, queues, trees, searching, sorting, and exceptions, with an emphasis on their implementation. 
%We collected the assignments and all available associated materials (e.g., reference solutions, test inputs, testing scripts, etc.) from the course instructors.
In total, we select a set of 58 programming assignments (30 from CS1 and 28 from CS2).
We discard 4 graphical user interface-based assignments from CS1, as creating test cases to check their correctness would require non-trivial efforts.
Next, we divide each assignment into multiple tasks, as one assignment can contain multiple problems, and categorize them into two types:
{\em short problems}, which require students to implement a single clearly-specified
function or class; and
{\em long problems}, which are more complex and which either require students to implement
multiple functions or classes that depend on each other, or else leave the required number of
functions or classes unspecified.
Our corpus contains a total of 84 short problems (20 from CS1 and 64 from CS2)
and 22 long problems (10 from CS1 and 12 from CS2). Examples of short and long problems are shown in Figure ~\ref{fig:short_and_long_problem} in Appendix~\ref{appen:short:long}.
We decide not to select any programming assignments from an open dataset for several reasons. Firstly, the evaluation of open datasets might hinder the generalizability of our findings, e.g., performance on open datasets might significantly vary from the closed one where problems were curated from the ground up. Secondly, to evaluate the proposed approaches using our methodology, it is essential to have problems with accurate and reliable solutions and test cases to grade them accurately. However, we did not find any such datasets that meet this requirement.
      
\noindent
\textbf{Creating Test Oracle.}
We create \textit{test oracles} to check \textit{correctness scores} of a given assignment solution. Given a solution, a test Oracle script runs a predefined set of test cases and outputs the percentage of test cases passed by the solution. To build these scripts, we reuse the test cases obtained from the instructor. We form two groups among the authors of this paper to create and validate these test oracles. One group creates the scripts for a selected assignment set, and another validates them.

%% mistral-large-2402
%% CodeLlama 7B
\noindent
\textbf{Model Selection.}
We consider five LLMs for this study: \GPT{}~\cite{gpt3.5}, \GITHUBCOPILOT{}~\cite{github_copilot}, \Mistral{}~\cite{mistral}, \CLLaMA{}~\cite{DBLP:journals/corr/abs-2308-12950} and \CODERL{}~\cite{CodeRL}. \GPT{} is used behind ChatGPT, and Mistral-Large is used behind Mistral AI chat. \GITHUBCOPILOT{} is an IDE (e.g., JetBrains IDEs, Visual Studio, etc.) plugin developed by GitHub that is powered by OpenAI's Codex model. We select these five models for their availability to \textit{fresh} CS students. We included \CLLaMA{} and \CODERL{} for their wide accessibility. 
%We evaluate \GPT{}, \GITHUBCOPILOT{}, \Mistral{} and \CLLaMA{}'s performance in a black-box setting. As comprehensive experimentation in a black-box setting is costly, we choose \CODERL{} as a \textit{whitebox surrogate} to guide our perturbation in Step 2 as specified in Figure~\ref{fig:overview:framework} (details in Section~\ref{sec:step2}).
The details of our code generation methods and the model versions and parameters are described in Appendix~\ref{app:methodology}; The most important point here is that we set any relevant parameters to values that produce the best possible solutions, upload the problem prompt into the LLM, and evaluate the solutions generated.
% For \CODERL{} this entails setting the \textit{temperature} to $0$ and the output token limit to its maximum allowable limit; for \GITHUBCOPILOT{}, we generate multiple solutions for each prompt and use our test oracles to select the one that has the highest score; for \GPT{}, we set the \textit{temperature} to 0 to obtain the best solution deterministically. Since \GPT{} is a general-purpose language model not specifically designed for code generation, we add a sentence to the problem statement instructing it to omit explanations and produce only code.

\subsection{Results: LLM performance}\label{step1:result}

We use all the short (84) and long (22) problems to evaluate the performance of the LLMs considered in our assignment corpus. For a given set of assignments, we define an LLM's performance as the average correctness scores of the corresponding solutions it generates. We generate correctness scores (the portion of the test cases that pass) with our test oracles.

\noindent
\textbf{Performance on CS1 Problems.}
% \SR{@Rubin, @Alex, and @Saiful, is it the same for CodeLlaMA and Mistral? Please confirm.}
% \AC{Yes it is the same, Mistral and CodeLlaMa struggle with CS1 problems as well.}
The LLMs we test do not generate completely correct solutions to any of the problems in our CS1 problem set. For two short and 5 long problems, \GPT{} refuses to generate any solutions due to triggering academic integrity safeguards. We discuss other possible reasons for this somewhat surprising result in Section~\ref{sec:rq1:discussion}.

% Please add the following required packages to your document preamble:
% \usepackage{multirow}
% \usepackage{graphicx}
\begin{table}[h]
\centering
\caption{LLMs' performance on CS2 problems.}
\label{tab:rq1_performance}
\resizebox{\columnwidth}{!}{%
\begin{tabular}{|c|ccc|ccc|}
\hline

\multirow{3}{*}{\textbf{Model}} 
& \multicolumn{3}{c|}{\textbf{Short (64)}}                
& \multicolumn{3}{c|}{\textbf{Long (12)}}                   \\ \cline{2-7}

& \multicolumn{1}{c|}{\multirow{2}{*}{Mean}}  
& \multicolumn{1}{c|}{Min} & Max & \multicolumn{1}{c|}{Mean} & \multicolumn{1}{c|}{Min} & Max \\ 

& \multicolumn{1}{c|}{}
& \multicolumn{1}{c|}{(Count)}
& \multicolumn{1}{c|}{(Count)}
& \multicolumn{1}{c|}{}
& \multicolumn{1}{c|}{(Count)}
& \multicolumn{1}{c|}{(Count)} \\ \hline

CodeRL & \multicolumn{1}{c|}{12.47} & \multicolumn{1}{c|}{0 (48)}   & 100 (3) & \multicolumn{1}{c|}{0.0}    & \multicolumn{1}{c|}{0 (12)}   & 0 (12)  \\ \hline

Code Llama & \multicolumn{1}{c|}{16.07} & \multicolumn{1}{c|}{0 (49)} & 100 (5) & \multicolumn{1}{c|}{0.83} & \multicolumn{1}{c|}{0 (11)} & 100 (1) \\ \hline

Mistral & \multicolumn{1}{c|}{50.09} & \multicolumn{1}{c|}{0 (26)} & 100 (23) & \multicolumn{1}{c|}{25.31} & \multicolumn{1}{c|}{0 (7)} & 100 (1) \\ \hline

GPT-3.5        & \multicolumn{1}{c|}{41.60} & \multicolumn{1}{c|}{0 (30)} & 100 (17) & \multicolumn{1}{c|}{8.33}  & \multicolumn{1}{c|}{0 (11)} & 100 (1) \\ \hline
GitHub Copilot & \multicolumn{1}{c|}{51.47} & \multicolumn{1}{c|}{0 (26)} & 100 (24) & \multicolumn{1}{c|}{26.99} & \multicolumn{1}{c|}{0 (6)} & 100 (2) \\ \hline
\end{tabular}%
}
\end{table}

\noindent
\textbf{Performance on CS2 Problems.}
The performance of the LLMs on our CS2 problem set is shown in Table \ref{tab:rq1_performance}.
By and large, they perform better than on the CS1 problems. 
\CODERL{} has the worst performance of the five LLMs tested: while it can construct correct solutions for some of the short problems with an average score of 12.5\% for the short problems, it fails to solve any of the long problems.
\GPT{} does somewhat better, scoring 41.6\% for the short problems and 8.3\% for the long
problems. While \Mistral{}'s performance was closer, \GITHUBCOPILOT{} had the best performance, with an average score of 51.5\% for the short problems and 27\% for the long problems.
\SUMMARY{\textbf{Finding 1:} All five LLMs fail to solve CS1 problems. For CS2, \GITHUBCOPILOT{} performed best, with an average score of 51.5\% for short and 27\% for long assignments.}

\subsection{Discussion on the Findings}\label{sec:rq1:discussion}
The LLMs' lack of success with CS1 problems is unexpected. Possible reasons for this include:
(1) %\sout{many of them are simple problems unlikely to be of sufficient general interest to show up in code repositories and thereby appear in LLM training sets;}
many of them are very specific problems unlikely to be of sufficient general interest to show up in code repositories and thereby appear in LLM training sets, providing a challenge for the LLMs to match the output required by the test oracles exactly;
(2) information relevant to some of the problems is provided graphically (60\% CS1 problems), sometimes in the form of ASCII art (Figure~\ref{fig:cs1:problem}), which was difficult for the LLMs to process; and
(3) assignments are often very specific regarding names of input/output files, classes, methods, etc., and the LLMs had trouble matching these specifics.
%%We note in passing that the second and third items may offer avenues for hardening programming assignments against LLM-assisted cheating in a way that is orthogonal to the prompt perturbation techniques discussed later in this paper.
These results are at odds with other research that suggests that LLMs can be effective in solving introductory programming problems \cite{DBLP:conf/ace/Finnie-AnsleyDB22,DBLP:conf/ace/Finnie-AnsleyDL23}.  Possible reasons for this difference include: (1) differences in the problems used in different studies, given that there is no consensus on what the specific content of CS1 and CS2 courses ought to be \cite{Hertz-SIGCSE2010}; and (2) methodological differences between studies, e.g.,  Finnie-Ansley {\em et al.} manually repaired minor errors in the LLM-generated solutions \cite{DBLP:conf/ace/Finnie-AnsleyDB22} while we did not.
Although the LLMs do not generate correct solutions for any of the CS1 problems, in some cases, they generate code that is \textit{close to correct} and could potentially be massaged to a correct solution by a student.

%\SR{@All, can it be the case that previous studies considered close to correct solutions as correct? Are we the only ones who used test cases to evaluate the correctness?} 
%\SR{What else can we say to prove the legitimacy of our results in comparison with SOTA. Reviewers (Including the meta-reviewer) raised concerns about this.}
%\AC{I rememeber one paper I read did manually fix small errors in the solutions generated by the LLMs when benchmarking, but they were dealing with a much smaller dataset than we were.}\SR{Great point! Can you please dig the papers so that we can cite them? Also, the exact number of assignments they used (if possible?) -- That will be awesome!!}
%\AC{The paper is "The Robots Are Coming: Exploring the Implications of OpenAI Codex on Introductory Programming" (https://dl.acm.org/doi/abs/10.1145/3511861.3511863) and they say they have 23 programming problems in the beginning of section 3.}
%\AC{I'm not sure where exactly we should cite the paper though, so I'm leaving it in the comment for now.}
% \SD{@Alex: I added in some text about this study above, please take a look and check that I'm being truthful.}
% We explore these issues further in the user study discussed in Section~\ref{sec:rq3:user-study}.

For the CS2 problems, there is a noticeable difference between LLM performance on short problems, which involve creating a single clearly specified function or class, and long problems, which are more complex and involve interactions between multiple functions or classes.
All of the LLMs generate correct solutions for some short problems but fail to generate correct solutions for others; 
while \CODERL{} fails to generate any correct solutions for any of the long problems. While \CLLaMA{} struggled too -- \GPT, \Mistral{} and
\GITHUBCOPILOT{} were able to generate correct solutions for some of the long problems.
Once again, for some of the problems, the LLM-generated code is close to correct, and students could potentially massage them manually into working solutions.  
% \SUMMARY{\textbf{Finding 2:} @Saiful, Here goes the summary of the discussion insights (Highlight what problems are more prone to LLM-assisted cheating).}

\section{Exploring Perturbations (Step 2)}\label{sec:step2} 
In this section, we explore the following research question: 
\textit{How can we leverage black-box adversarial perturbation techniques to impede LLM-assisted solution generation?} Towards that end, following existing literature, we design several perturbation techniques and measure their efficacy on the assignments that LLMs solved with non-zero correctness scores. For a given perturbation technique, we define its efficacy as follows.

\begin{definition}[Efficacy]\label{efficacy} 
    The efficacy of a perturbation technique for a given assignment is the reduction of the LLM's correctness score from the base correctness score on the assignment. % Here, the base correctness score indicates an LLM's correctness score with no perturbation technique applied.
% \resizebox{\columnwidth}{!}{%
% \SR{Rewrite it as: $max\{0,~BS - PS\}$?}
\begin{align*}
\text{Efficacy} &= max\Bigg\{0,~100\times\frac{S_{no\_prtrb} - S_{prtrb}}{S_{no\_prtrb}}\Bigg\} \\
\text{where, }&\\
S_{no\_prtrb} &= \text{Correctness with no perturbation}\\
S_{prtrb} &= \text{Correctness with perturbation} \\
% \delta &= \text{Passing score}
\end{align*}
 % }
\end{definition}

Given the same amount of drops in the correctness score, our efficacy favors the lower correctness score after perturbation. This is because, for example, a drop of 30\% from 70\% is more favorable than a drop of 30\% from 100\%, as the former has a more drastic impact on the overall grade.

\subsection{Perturbation Methodology}

We design \underline{ten perturbation techniques} under two broad categories, \textit{core} and \textit{exploratory}.

\noindent
\textbf{Core perturbations.} Under this category, we design seven principled techniques with four end-to-end automated perturbation strategies, \textit{i)} synonym substitution, \textit{ii)} rephrasing sentences, \textit{iii)} replacing characters with \textit{Unicode} lookalikes, and \textit{iv)} removing contents. We apply these strategies to different perturbation units, i.e., characters, tokens, words, and sentences. Perturbation units indicate the unit of changes we make at once. Inspired by explainability-guided adversarial sample generation literature~\cite{DBLP:conf/sigsoft/SunXTDLWZCN23, DBLP:conf/ijcnn/RosenbergMBGSD20}, we use SHAP (SHapley Additive exPlanations)~\cite{SHAP_LundbergL17} with \CODERL{} as the \textit{surrogate} model to select candidate units for perturbations. Specifically, we use Shapley values to compute the top-ranked tokens for perturbation. %For example, given a perturbation strategy, we replace the top 5 tokens to create variants of \textit{perturbed} assignment statements. 
For example, for \textit{Character (remove)} perturbation, we remove a random character from each token to generate one variant; for \textit{Token (remove)} perturbation, we remove all 5 tokens to generate one variant, and for the synonym morphs, we may have many synonyms for one token, and generate many variants. For \textit{Token (unicode)} perturbation, we replace all 5 tokens with Unicode characters to generate one variant. For example, we replaced \textit{a}, \textit{c}, and \textit{y} with \textit{à}, \textit{ċ}, and \textit{ý}, respectively. We use the token rank for all the other perturbation units except for sentences. We rank the sentences by accumulating the Shapley values of the tokens corresponding to a given sentence for sentence perturbations. We add a detailed description of each technique in the Appendix~\ref{app:perturb}.

% Please add the following required packages to your document preamble:
% \usepackage{graphicx}
\begin{table*}[!ht]
\centering
\caption{Average efficacy of the perturbation techniques. All the perturbations combined caused performance degradation for a significant portion of assignments, which was dictated by ``\texttt{Sentence (remove)}'' and ``\texttt{Prompt (unicode)}'' perturbations.}
% \SR{In this version, our efficacy looks unfairly bad. Let's also include the base scores in this table? Let's order the Models as follows CodeRL, CodeLlama, Mistral, GPT-3.5 and GitHub Copilot.} \RY{What are the "Problem Count"s? It seems CodeRL and CodeLlama results are identical as well.}\SIS{Thanks for noticing. Fixed.}}
\label{tab:rq2_efficacy}
\resizebox{\textwidth}{!}{%
\begin{tabular}{|l|cc|cc|cc|cc|cc|}
\hline
 &
  \multicolumn{2}{c|}{\textbf{CodeRL}} &
  \multicolumn{2}{c|}{\textbf{Code Llama}} &
  \multicolumn{2}{c|}{\textbf{Mistral}} &
  \multicolumn{2}{c|}{\textbf{GPT-3.5}} &
  \multicolumn{2}{c|}{\textbf{GitHub Copilot}} \\ \hline
% \textbf{Base Score (Avg.) $\rightarrow$} &
%   \multicolumn{2}{c|}{10.5} &
%   \multicolumn{2}{c|}{13.66} &
%   \multicolumn{2}{c|}{46.18} &
%   \multicolumn{2}{c|}{36.35} &
%   \multicolumn{2}{c|}{47.6} \\ \hline \hline
\textbf{Perturbations} &
  \multicolumn{1}{c|}{\begin{tabular}[c]{@{}c@{}}Problem\\ Count (\%)\end{tabular}} &
  \begin{tabular}[c]{@{}c@{}}Avg. \\ Efficacy\end{tabular} &
  \multicolumn{1}{c|}{\begin{tabular}[c]{@{}c@{}}Problem\\ Count (\%)\end{tabular}} &
  \begin{tabular}[c]{@{}c@{}}Avg.\\ Efficacy\end{tabular} &
  \multicolumn{1}{c|}{\begin{tabular}[c]{@{}c@{}}Problem\\ Count (\%)\end{tabular}} &
  \begin{tabular}[c]{@{}c@{}}Avg.\\ Efficacy\end{tabular} &
  \multicolumn{1}{c|}{\begin{tabular}[c]{@{}c@{}}Problem\\ Count (\%)\end{tabular}} &
  \begin{tabular}[c]{@{}c@{}}Avg.\\ Efficacy\end{tabular} &
  \multicolumn{1}{c|}{\begin{tabular}[c]{@{}c@{}}Problem\\ Count (\%)\end{tabular}} &
  \begin{tabular}[c]{@{}c@{}}Avg.\\ Efficacy\end{tabular} \\ \hline
\texttt{Character (remove)} &
  \multicolumn{1}{c|}{31.25} &
  7.81 &
  \multicolumn{1}{c|}{50.0} &
  12.19 &
  \multicolumn{1}{c|}{32.56} &
  24.03 &
  \multicolumn{1}{c|}{40.0} &
  22.4 &
  \multicolumn{1}{c|}{25.0} &
  25.17 \\ \hline
\texttt{Token (unicode)} &
  \multicolumn{1}{c|}{43.75} &
  10.94 &
  \multicolumn{1}{c|}{50.0} &
  12.5 &
  \multicolumn{1}{c|}{20.93} &
  25.27 &
  \multicolumn{1}{c|}{34.29} &
  18.49 &
  \multicolumn{1}{c|}{11.36} &
  14.78 \\ \hline
\texttt{Token (remove)} &
  \multicolumn{1}{c|}{25.0} &
  6.25 &
  \multicolumn{1}{c|}{56.25} &
  20.61 &
  \multicolumn{1}{c|}{20.93} &
  18.07 &
  \multicolumn{1}{c|}{37.14} &
  17.84 &
  \multicolumn{1}{c|}{34.09} &
  43.79 \\ \hline
\texttt{Token (synonym)} &
  \multicolumn{1}{c|}{56.25} &
  7.65 &
  \multicolumn{1}{c|}{81.25} &
  16.57 &
  \multicolumn{1}{c|}{39.53} &
  30.56 &
  \multicolumn{1}{c|}{42.86} &
  23.81 &
  \multicolumn{1}{c|}{38.64} &
  26.83 \\ \hline
\texttt{Tokens (synonym)} &
  \multicolumn{1}{c|}{56.25} &
  9.17 &
  \multicolumn{1}{c|}{87.5} &
  17.73 &
  \multicolumn{1}{c|}{44.19} &
  29.25 &
  \multicolumn{1}{c|}{45.71} &
  20.95 &
  \multicolumn{1}{c|}{34.09} &
  35.1 \\ \hline
\texttt{Sentences (rephrase)} &
  \multicolumn{1}{c|}{75.0} &
  11.85 &
  \multicolumn{1}{c|}{87.5} &
  18.05 &
  \multicolumn{1}{c|}{23.26} &
  9.28 &
  \multicolumn{1}{c|}{51.43} &
  17.36 &
  \multicolumn{1}{c|}{22.73} &
  21.92 \\ \hline
\texttt{Sentences (remove)} &
  \multicolumn{1}{c|}{\textbf{93.75}} &
  14.07 &
  \multicolumn{1}{c|}{68.75} &
  15.64 &
  \multicolumn{1}{c|}{\textbf{90.7}} &
  42.98 &
  \multicolumn{1}{c|}{\textbf{88.57}} &
  30.71 &
  \multicolumn{1}{c|}{\textbf{79.55}} &
  \textbf{60.94} \\ \hline
\texttt{Prompt (unicode)} &
  \multicolumn{1}{c|}{\textbf{93.75}} &
  \textbf{23.44} &
  \multicolumn{1}{c|}{\textbf{100}} &
  \textbf{31.77} &
  \multicolumn{1}{c|}{79.07} &
  \textbf{86.2} &
  \multicolumn{1}{c|}{54.29} &
  \textbf{33.23} &
  \multicolumn{1}{c|}{43.18} &
  47.36 \\ \hline
\texttt{Random (insert)} &
  \multicolumn{1}{c|}{6.25} &
  1.56 &
  \multicolumn{1}{c|}{50} &
  17.71 &
  \multicolumn{1}{c|}{0.0} &
  0.0 &
  \multicolumn{1}{c|}{11.43} &
  5.47 &
  \multicolumn{1}{c|}{15.9} &
  17.32 \\ \hline
\texttt{Random (replace)} &
  \multicolumn{1}{c|}{37.5} &
  9.11 &
  \multicolumn{1}{c|}{\textbf{100}} &
  \textbf{31.77} &
  \multicolumn{1}{c|}{\textbf{90.7}} &
  87.86 &
  \multicolumn{1}{c|}{25.71} &
  18.68 &
  \multicolumn{1}{c|}{13.64} &
  9.11 \\ \hline
\texttt{Combined} &
  \multicolumn{1}{c|}{\textbf{93.75}} &
  \textbf{100} &
  \multicolumn{1}{c|}{\textbf{100}} &
  \textbf{100} &
  \multicolumn{1}{c|}{\textbf{100}} &
  \textbf{100} &
  \multicolumn{1}{c|}{\textbf{97.14}} &
  \textbf{91.21} &
  \multicolumn{1}{c|}{\textbf{90.91}} &
  \textbf{80.03} \\ \hline
\end{tabular}%
}
\end{table*}

\noindent
\textbf{Exploratory perturbations.} We design three additional techniques to explore the potential of two different insights. For example, existing studies show evidence that LLMs are prone to memorizing training data~\cite{DBLP:journals/corr/abs-2112-12938, DBLP:conf/uss/CarliniTWJHLRBS21, DBLP:conf/iclr/CarliniIJLTZ23}. Thus, these models are highly sensitive to input variations~\cite{DBLP:conf/iclr/ZhangLCDBTHC22, DBLP:conf/acl/Jin0SC022, DBLP:conf/chi/ReynoldsM21}. Under this hypothesis, replacing specific tokens with random strings may significantly influence performance, as such substitution may alter the context~\cite{DBLP:conf/icml/ShiCMSDCSZ23, DBLP:journals/corr/abs-2307-03172, DBLP:conf/nips/WangXWG0GA021}. We design a new exploratory perturbation technique to leverage this insight. Under this technique, we tweak assignments by replacing file names, function names, and class names specified in the problem statement with random words, where these names are discovered manually. Another example is that to understand the resiliency of LLMs on Unicode lookalikes~\cite{ShettySF18, BoucherS0P22}, we create a mechanism to replace all possible characters with Unicode lookalikes in the \textit{entire} assignment statement.

\begin{figure}[ht]
    \centering
    \includegraphics[width=\linewidth]{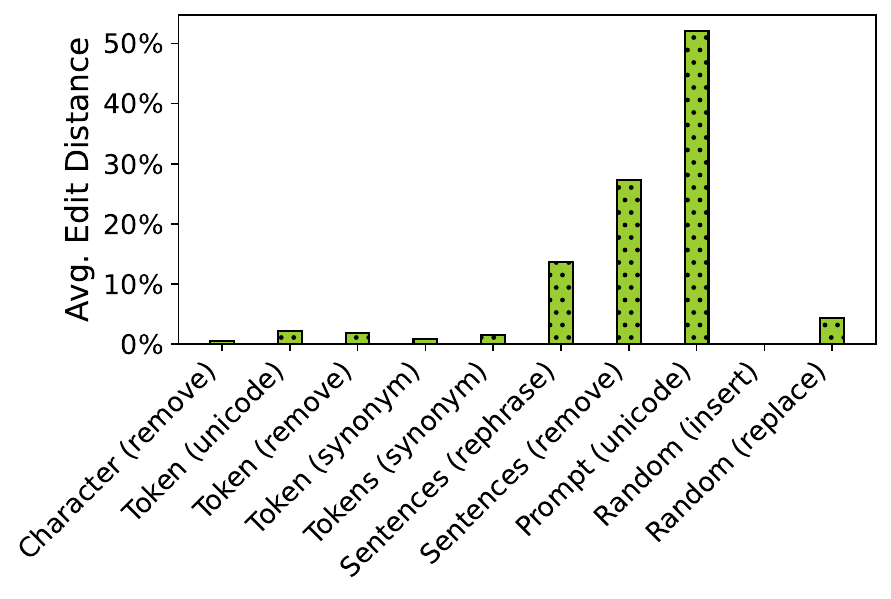}
    \vspace{-20pt}
    \caption{The average changes caused by the perturbation techniques are calculated as the edit distance between the original and the perturbed assignments.}
    
    \label{fig:rq2_edit_distance}
\end{figure}

\subsection{Results: Perturbation Performance}

We measure the performance of our perturbation techniques on the assignments that LLMs solved with non-zero correctness scores. % Specifically, we seek to answer the following two questions: \textbf{(RQ1)} \textit{What was the efficacy of each perturbation technique in reducing LLM performance?} \textbf{(RQ2)} \textit{How many average changes do they cause on the assignment descriptions?}

\noindent
\textbf{Perturbation Efficacy.} Table \ref{tab:rq2_efficacy} depicts the efficacy of all our perturbations. All the perturbations combined cause performance degradation in all five models for most of the assignments we tested. Combined perturbation efficacy is the average efficacy of the best perturbation technique for each problem, i.e.,

\vspace{-20pt}
\[
\text{Combined Efficacy} = \frac{1}{n} \sum_{i=1}^n \max \{E_{i}\}, \text{where,}
\]
\begin{itemize}
    \item $n$ is the total number of problems,
    \item $E_{i}$ is the list of efficacy scores of all the perturbation techniques on the $i$-th problem
\end{itemize}
% \SR{@Saiful/@Rubin, please write a sentence on how did we combine the results.}
% \RY{Above in blue is the full definition}
The performance is mostly dictated by ``remove sentence'' and followed by ``assignment-wide substitution with Unicodes'' perturbations. However, the average \textit{edit distance} for these two techniques is much higher, making them riskier for \textit{detection} (Figure~\ref{fig:rq2_edit_distance}), which we discuss next. 

\noindent
\textbf{Changes in the original prompt.} A higher proportion of changes caused by a perturbation technique risks both understandability and detectability.
We use the edit distance between the original and perturbed assignment statements to quantify the changes for a given perturbation technique. Note that edit distance is not the ideal method to capture the drifts (if any) caused by Unicode replacements (visual) and synonyms (conceptual); However, it gives a picture of how much the perturbed prompt was altered from the original one.
Figure \ref{fig:rq2_edit_distance} depicts the average edit distance of the perturbation techniques on the assignments with positive efficacy (i.e., causing performance degradation). Except for sentence and prompt-wide perturbations, all the other techniques require a small (<5\%) amount of perturbation to the problem statements. This is because they are performed on a small portion of characters or tokens, making them less expensive.
\SUMMARY{\textbf{Finding 2:} The combination of all the perturbations covers more than 90\% of the problems with efficacy >80\% for all five models. \textit{High-change} perturbations have high efficacy.}

\noindent
\textbf{Why perturbations failed?} To understand why our perturbation techniques may have failed, we study the two sets of assignments where they succeeded and failed. Under the \underline{succeeded} category, we select assignments where the average efficacy was high (greater than 90) for at least half of the perturbation techniques. For \underline{failed} category, we select assignments with efficacy 0 for all the techniques. Next, we randomly select 10 samples for each category and study the \textit{variety} in the generated solutions by the LLMs under various perturbation techniques. For a given assignment, we measure variety by directly comparing all the solutions and counting unique variations. We observe that the average number of unique variations per problem is 13.9 and 26.0 for problems where perturbation failed and succeeded, respectively. % We note that as perturbations fail in cases where they fail to trigger variety in the solution space, instructors can avoid such assignments to harden the solution generation.
To determine the uniqueness of solutions, we use AST similarity. Comparison of the ASTs of the codes that are the same except for different variable names gets a similarity score of 100, and formatting differences between solutions will be ignored. We use a threshold of 90 when determining if a program is unique.

\SUMMARY{\textbf{Finding 3:} High variations in generated solutions strongly correlate with high success rates for a given perturbation technique.}

\section{Field Experiment (Step  3)}\label{sec:rq3:user-study}
In this step, we aim to understand how students would detect and reverse our perturbations. This would provide valuable insights into the potential of the perturbation techniques for impeding actual LLM-assisted cheating.

% Toward that, we run an IRB-approved user study-based field experiment on undergraduate students, which we discuss next.
% \SR{Address the Meta-reviewer's concern about the user study, which we discuss next.}

% Our goal is to measure the effectiveness of our morphing techniques to harden LLM-assisted cheating in a real environment. To achieve this goal, we analyze how difficult it is for the students to solve the assignment problems in the presence of different morphing techniques with the help of LLMs. This section explains the methodology of the user study and our findings from the user study.

\subsection{Methodology}

\noindent
\textbf{User Study Design.}
We recruited 30 undergraduate students who had previously completed CS1 and CS2 courses from the same university to participate in this IRB-approved user study. Each participant was awarded \$20 for their participation. During this study, each student was explicitly asked to use ChatGPT to solve 3 assignments over one week and submit the entire chat history in a post-study survey. After the experimentation, we asked the participants to submit their chat history with ChatGPT and observed that all of the participants used ChatGPT-3.5, except for one who used the ChatGPT-4.0 version. We discarded the data from that user.

The details of specific instructions to the students are added in Appendix \ref{sec:instructions_to_the_participants}. We assign each assignment-perturbation pair to at least three participants to cover redundancy and diversity. \textit{This includes no perturbation cases, too, which indicates the base performance.}
Our post-study survey also asks whether students noticed anything ``unusual'' in the assignment description, how they validated solutions, etc. (details in Table~\ref{tab:userstudy:questions}). 
 Note that for ethical reasons, we chose to run the study on students who already took the courses (Demographic information in Table~\ref{tab:participant_demography}). We discuss its impact on the outcome in Section~\ref{limitations}.
 
 % Since the goal of the study is to understand how students would detect and reverse the perturbation, we think this 

%% We do this for ethical reasons. Note that, as the students are already exposed to the problems, our study results would underline a conservative estimation of the true potential of perturbation-based hardening in LLM-based solution generation.

\textbf{Problem Selection.} For this study, we select assignments for which the efficacy score for at least one perturbation was 80 on \GPT{}, which powers ChatGPT. We chose 6 assignments with at least 3 perturbed versions, from this initial list, under 3 different techniques. Table \ref{tab:problems_user_study} shows the problem and perturbation technique pairs selected for the user study. \texttt{Prompt (Original)} indicates prompt with no perturbation. 
%\sout{We handle the removal of content-based (i.e., characters, tokens, etc.) perturbations in the user study by replacing them with images so that they stay \textit{removed} in straightforward copy attempts.} 
We recognize that removal of content (i.e., characters, tokens, etc.) from the assignment text will be easily detected by students. To remedy this, we replace the removed content with images of the characters that were removed in an attempt to make the text look as visually identical to the original assignment as possible. We assume that students will copy and paste the text from the assignment into the ChatGPT input box, and because images do not get copied, the text pasted into ChatGPT will be pertubed. Table~\ref{tab:prob_morph_dist} in Appendix~\ref{app:userstudy} shows the distributions of the number of participants for different variants of the assignments.

\begin{table}[ht]
\caption{Selected assignments and corresponding perturbation techniques for the user study. \texttt{Prompt (Original)} indicates prompt with no perturbation.}
\label{tab:problems_user_study}
\centering
\resizebox{0.9\columnwidth}{!}{
\begin{tabular}{|l|cccccc|}
\hline
\multirow{2}{*}{\textbf{Perturbations}} & \multicolumn{6}{c|}{\textbf{Assignments}}                                                                                                        \\ \cline{2-7} 
                           & \multicolumn{1}{c|}{\#1} & \multicolumn{1}{c|}{\#2} & \multicolumn{1}{c|}{\#3} & \multicolumn{1}{c|}{\#4} & \multicolumn{1}{c|}{\#5} & \#6 \\ \hline
\texttt{Prompt (original)}   & \multicolumn{1}{c|}{\checkmark}  & \multicolumn{1}{c|}{\checkmark}  & \multicolumn{1}{c|}{\checkmark}  & \multicolumn{1}{c|}{\checkmark}  & \multicolumn{1}{c|}{\checkmark}  & \checkmark  \\ \hline
\texttt{Character (remove)}   & \multicolumn{1}{c|}{-}  & \multicolumn{1}{c|}{\checkmark}  & \multicolumn{1}{c|}{-}  & \multicolumn{1}{c|}{-}  & \multicolumn{1}{c|}{-}  & \checkmark  \\ \hline
\texttt{Token (unicode)}  & \multicolumn{1}{c|}{\checkmark}  & \multicolumn{1}{c|}{\checkmark}  & \multicolumn{1}{c|}{\checkmark}  & \multicolumn{1}{c|}{-}  & \multicolumn{1}{c|}{-}  & \checkmark  \\ \hline
\texttt{Tokens (remove)}      & \multicolumn{1}{c|}{\checkmark}  & \multicolumn{1}{c|}{-}  & \multicolumn{1}{c|}{-}  & \multicolumn{1}{c|}{-}  & \multicolumn{1}{c|}{\checkmark}  & -  \\ \hline
\texttt{Sentences (rephrase)} & \multicolumn{1}{c|}{\checkmark}  & \multicolumn{1}{c|}{-}  & \multicolumn{1}{c|}{-}  & \multicolumn{1}{c|}{-}  & \multicolumn{1}{c|}{-}  & -  \\ \hline
\texttt{Sentences (remove)}   & \multicolumn{1}{c|}{\checkmark}  & \multicolumn{1}{c|}{\checkmark}  & \multicolumn{1}{c|}{-}  & \multicolumn{1}{c|}{\checkmark}  & \multicolumn{1}{c|}{-}  & -  \\ \hline
\texttt{Prompt (unicode)} & \multicolumn{1}{c|}{\checkmark}  & \multicolumn{1}{c|}{-}  & \multicolumn{1}{c|}{\checkmark}  & \multicolumn{1}{c|}{\checkmark}  & \multicolumn{1}{c|}{\checkmark}  & \checkmark  \\ \hline
\texttt{Random (replace)}     & \multicolumn{1}{c|}{\checkmark}  & \multicolumn{1}{c|}{\checkmark}  & \multicolumn{1}{c|}{\checkmark}  & \multicolumn{1}{c|}{-}  & \multicolumn{1}{c|}{-}  & -  \\ \hline
\end{tabular}%
}
\end{table}

%In our study, first, we prepare the study materials, which include the problems and the survey form. Then, we recruit the study participants and conduct the user study. We adopt a mixed design analysis~\cite{Heigham2009}\cite{Quaium2023}\cite{McKim2017}, where we perform both qualitative and quantitative analysis. For the qualitative analysis, we performed the thematic analysis of the responses. We perform quantitative analysis to measure the correctness of the participants' provided solutions.

\noindent
\textbf{Analyzing the textual Responses.} Answers to some of the questions in our post-study questionnaire were open-ended. Thus, to systematically analyze those responses, we use thematic analysis, where the goal is to identify the concepts (known as \textit{codebook}) and organize them under different themes~\cite{JasonThematicAnalysis2015, Quaium2023}. Two authors participate in the process to avoid human bias. Our thematic analysis found that students use 5 different approaches to neutralize perturbations and 11 different approaches to validate LLM-generated solutions. We present a detailed description of the method and the codebook in the Appendix~\ref{app:userstudy}. 

\noindent
\textbf{Analyzing Solutions.}
The performance of black-box models changes over time. Without taking this into account, one might come to erroneous conclusions. For example, Figure~\ref{fig:rq2_model_checkpoint_average_score_barplot} shows the performance of different model checkpoints on the assignment statements we use for the user study since we computed the efficacy with model checkpoint 0301. However, to ensure consistency in calculating the efficacy of the perturbation techniques in impeding the actual cheating, one needs to calculate the correctness scores for both the perturbed and unperturbed versions of the assignments on the \textit{same} model checkpoints. Thus, we use the average correctness scores of unperturbed assignments to compute the average efficacy of a given perturbation technique.

\subsection{Analysis Results}
% \SR{Rewrite the questions to address meta reviewer's concern.}
In this section, we present the results of our field experiment to answer the following three questions: \textbf{Q1:} \textit{How effective are the perturbations, in general, in impeding LLM-assisted solution generation?} \textbf{Q2:} \textit{How does the detectability affect efficacy?} and \textbf{Q3:} \textit{What techniques do students adopt to avoid perturbations, and how do they validate their generated solutions?}

\begin{table}[ht]
\centering
\caption{Efficacy for each perturbation technique on the 6 problems we used for the user study.}
\label{tab:rq3_efficacy_for_morphs_barplot}
\resizebox{0.7\columnwidth}{!}{%
\begin{tabular}{l|c}
\hline
\hline

\textbf{Perturbations}             & \textbf{Avg. Efficacy} \\ \hline \hline

\texttt{\textbf{No perturbation}}         & \textbf{71.28} (Base Score) \\ \hline

\texttt{Character (remove)}       & 6.67\%                     \\ \hline
\texttt{Token (unicode)}         & 18.08\%                    \\ \hline
\texttt{Token (Remove)}          & 0.0\%                      \\ \hline
\texttt{Sentence (Rephrase)}     & 0.0\%                      \\ \hline
\texttt{Sentences (Remove)}       & 10.0\%                    \\ \hline
\texttt{Prompt (unicode)}   & 31.25\%                     \\ \hline
\texttt{Random (Replace)}         & 15.91\%                     \\ \hline
\texttt{\textbf{Combined Results}}         & \textbf{76.67\%}                     \\ \hline

\end{tabular}
}
\end{table}

\noindent
\textbf{Impeding solution generation.} 
Overall, the perturbations are effective in impeding LLM-assisted solution generation. Although most of the perturbations have an efficacy lower than 32\%, in combination (selecting the best perturbation technique for each problem), their efficacy is around 77\%, where the base correctness score was 71.28 (Table \ref{tab:rq3_efficacy_for_morphs_barplot}). This means perturbation techniques reduced 77\% of the base score -- showing promise in impeding LLM-assisted cheating. One interesting finding is that the \texttt{Prompt (unicode)} perturbation drops the models' performance significantly. While most students notice it and exercise several strategies, they fail to sidestep it.

% Please add the following required packages to your document preamble:
% \usepackage{graphicx}
\begin{table}[ht]
\centering
\caption{Comparison of average efficacy for the perturbation techniques based on whether they were detected or not. For \texttt{Token (remove)} and \texttt{Sentence (rephase)}, ChatGPT (GPT-3.5) generated correct solutions without any tweaks from the students.}
\label{tab:noticed_vs_efficacy}
\resizebox{0.9\columnwidth}{!}{%
\begin{tabular}{|l|c|c|}
\hline
\textbf{Perturbations} & \textbf{Noticed(\%)} & \textbf{Unnoticed(\%)} \\ \hline
\texttt{Character (remove)}      & 0.0                & 16.0                 \\ \hline
\texttt{Token (unicode)}         & 6.67             & 43.75              \\ \hline
\texttt{Token (remove)}          & 0.0                & 0.0                  \\ \hline
\texttt{Sentences (rephrase)}    & 0.0                & 0.0                  \\ \hline
\texttt{Sentences (remove)}      & 16.67            & 0.0                  \\ \hline
\texttt{Prompt (unicode)}        & 35.71            & 0.0                  \\ \hline
\texttt{Random (replace)}        & 10.71             & 25.0                  \\ \hline
Total        & 15             & 15.43                  \\ \hline
\end{tabular}%
}
\end{table}

\noindent
\textbf{Detectability vs. Efficacy.}
Broadly, participants notice \textit{unusualness} in the assignments for all the perturbations (Table~\ref{tab:unnoticed_ratios}). In Table~\ref{tab:noticed_vs_efficacy}, we show the difference in efficacy based on whether the students notice a perturbation or not. % \RY{Is this referring to something else besides Fig 8? The previous plot before the revision (rq3\_efficacy\_morphing\_left\_noticed\_right\_corrected.pdf) didn't show high percentage of noticed cases as far as I remember.}
Overall, the average efficacy dropped (15.43\% to 15\%) for detectability.
Prompt/assignment-wide substitutions with Unicode lookalikes that alter a large portion of the assignment are easily noticed (Table~\ref{tab:unnoticed_ratios}). Despite the higher risk of being noticed, it still managed to deceive the model.
Higher efficacies in noticed cases of perturbations, such as the removal of sentences and prompt-wide Unicode substitution, suggest that noticing the perturbation does not imply that students were able to reverse the changes, especially if reversing involves some degree of effort.
Subtle perturbations, i.e., substitutions of tokens and removal of characters, showed great potential in tricking both the LLM and students, as they show higher efficacy when undetected. 

%% Sazzadur commented this out, as the following statement is inaccurate.
% The models are already bad at solving problems in cases of token removal and sentence rephrasing perturbations, which is also reflected in the user study with efficacy being 0. %\SR{@Rubin, We need to check the ``Token (remove)'' and ``Sentence (rephrase)'' cases. Efficacy 0 means what? Needs further investigation and explanation. In theory, without perturbation, the base score should be non-zero.}

% \begin{table}[htbp]
% \centering
% \caption{Unnoticed Ratios Across Perturbation Techniques}
% \label{tab:unnoticed_ratios}
% \resizebox{0.7\columnwidth}{!}{%
% \begin{tabular}{l|c}
% \hline
% \textbf{Perturbations}             & \textbf{Unnoticed} \\ \hline
% \texttt{Character (remove)}       & 5/12                     \\ \hline
% \texttt{Token (unicode)}      & 3/13                     \\ \hline
% \texttt{Token (Remove)}          & 2/7                      \\ \hline
% \texttt{Sentence (Rephrase)}     & 3/3                      \\ \hline
% \texttt{Sentences (Remove)}       & 4/10                     \\ \hline
% \texttt{Prompt (unicode)}   & 1/16                     \\ \hline
% \texttt{Random (Replace)}         & 5/11                     \\ \hline
% \end{tabular}
% }
% \end{table}

\begin{table}[htbp]
\centering
\caption{Unnoticed Ratios Across Perturbations}
\label{tab:unnoticed_ratios}
\resizebox{0.7\columnwidth}{!}{%
\begin{tabular}{l|c}
\hline
\textbf{Perturbations}             & \textbf{Unnoticed / Total} \\ \hline
\texttt{Character (remove)}       & 5/12                     \\ \hline
\texttt{Token (unicode)}      & 4/13                     \\ \hline
\texttt{Token (Remove)}          & 2/7                      \\ \hline
\texttt{Sentence (Rephrase)}     & 2/3                      \\ \hline
\texttt{Sentences (Remove)}       & 4/10                     \\ \hline
\texttt{Prompt (unicode)}   & 2/16                     \\ \hline
\texttt{Random (Replace)}         & 4/11                     \\ \hline
\end{tabular}
}
\end{table}

\SUMMARY{\textbf{Finding 4:} Subtle perturbations, i.e., substituting tokens or removing/replacing characters, when unnoticed, are likely to retain high efficacy in impeding actual cheating.}
\SUMMARY{\textbf{Finding 5:} The \textit{detectability} of a \textit{high-change} perturbation might not imply \textit{reversion}.}

\noindent
\textbf{Handling perturbed assignments.} 
We learn from the post-user study questionnaire that even if students noticed perturbations, in most cases (32 out of 49), they rely on ChatGPT to bypass them (Figure~\ref{fig:rq3_q2_coding_vs_morphing}). Other strategies they adopt are updating the assignment statement, rewriting incorrect ChatGPT-generated solutions, or writing the missing portions. The average efficacy against each of the strategies is highest at 31.11\%  when students impose \textit{`Update problem statement'}, followed by \textit{`No unusualness found'} at 15.43\% and \textit{`Expected to be bypassed'} at 9.17\%. When students try \textit{`Rewrite incorrect/missing portion'}, the perturbation efficacy is reduced to 0.

\noindent
\textbf{Validation apporaches.}
Approaches to validate the generated solutions also play a crucial role in detecting and fixing accuracy degradation. Most students report that they reviewed the generated code (72 out of 90 cases) or ran the code with the given test cases (55 out of 90 cases). Several of them report writing new test cases, too. A heatmap diagram of the validation approaches is presented in Figure~\ref{fig:rq3_q1_coding_heatmap_per_user} in Appendix~\ref{app:userstudy}.

\section{Discussion}\label{sec:discussion}

\noindent
\textbf{Impact of Model Evolution on solving assignments.}
To understand how our results might be affected as LLMs evolve, we compared the capabilities of \GPT{} and \GPTFOUR{}. Table \ref{tab:GPT-3.5_vs_GPT-4.0} shows a comparison. It can be seen that \GPTFOUR{} does perform slightly better than \GPT{} on the CS2 problems, and while \GPTFOUR{} scored just over 12\% on long problems and almost 16\% on short problems for CS1, \GPT{} scored 0\% on both, so \GPTFOUR{} evidently has some advanced capabilities that \GPT{} lacks.

% Please add the following required packages to your document preamble:
% \usepackage{multirow}
% \usepackage{graphicx}
\begin{table}[ht]
\centering
\caption{Performance comparison of GPT-3.5 and GPT-4.0 models on the CS introductory problems}
\label{tab:GPT-3.5_vs_GPT-4.0}
\resizebox{\columnwidth}{!}{%
\begin{tabular}{|c|cc|cc|cc|}
\hline
\multirow{2}{*}{\textbf{Model}} &
  \multicolumn{2}{c|}{\textbf{CS1}} &
  \multicolumn{2}{c|}{\textbf{CS2}} &
  \multicolumn{2}{c|}{\textbf{\begin{tabular}[c]{@{}c@{}}Perturbed CS2\\ (Selected)\end{tabular}}} \\ \cline{2-7} 
                   & \multicolumn{1}{c|}{Short} & Long  & \multicolumn{1}{c|}{Short} & Long  & \multicolumn{1}{c|}{Short} & Long  \\ \hline
\texttt{gpt-3.5-turbo-0301} & \multicolumn{1}{c|}{0.0}   & 0.0   & \multicolumn{1}{c|}{49.36} & 16.67 & \multicolumn{1}{c|}{29.31} & 17.43 \\ \hline
\texttt{gpt-4-0613}         & \multicolumn{1}{c|}{15.71} & 13.11 & \multicolumn{1}{c|}{56.14} & 23.57 & \multicolumn{1}{c|}{39.23} & 15.72 \\ \hline
\end{tabular}%
}
\end{table}

\noindent
\textbf{Impact of Model Evolution on perturbations.}
We run \GPTFOUR{} on the prompts generated by some of the promising perturbation techniques from the user study, i.e., \texttt{Sentences (remove)}, \texttt{Token (unicode)}, and \texttt{Prompt (unicode)}. 
% \SR{@Alex, how did you come to this ``1,113'' figure?}
% \AC{That is how many prompts we have for those morphing techniques.}
Out of the 1,113 prompts compared, \GPTFOUR~ outscored \GPT~ on 281 problems, while \GPT{} outscored \GPTFOUR{} on 107 problems (Table~\ref{tab:GPT-3.5_vs_GPT-4.0}). We observe that \GPT{} has built-in safeguards for academic integrity violations. Surprisingly, \GPTFOUR{} seems to lack such safeguards. For example, \GPT~ refuses to solve 8 problems for triggering such safeguards, but \GPTFOUR~ refuses none. 
This finding is concerning because it suggests that \GPTFOUR{} could potentially be more amenable to misuse for LLM-assisted cheating.

\section{Related Work}\label{sec:related-work}

%%In the realm of academic integrity, the emergence of Large Language Models (LLMs) such as ChatGPT presents both challenges and opportunities. 
% Research on LLMs relevant to academic integrity includes their capabilities in educational problem-solving, their resilience against adversarial attacks, and the role of explainable AI.

\noindent
\textbf{LLMs in Educational Problem Solving.} 
%%Studies examining the proficiency of LLMs in coding problem-solving provide essential insights. 
Finnie-Ansley {\em et al.} found that OpenAI Codex produced high-quality solutions for a set of CS1 and CS2 programming problems \cite{DBLP:conf/ace/Finnie-AnsleyDB22,DBLP:conf/ace/Finnie-AnsleyDL23}.
This suggests that LLM-assisted cheating in introductory programming courses has the potential to be problematic.
Other studies note that LLM-generated code can be of variable quality and sensitive to small changes to the prompt; this hints at the idea that tweaking the problem prompt can affect the usefulness of LLM-generated solutions for academic dishonesty.
For example, Wermelinger observes that ``Sometimes Copilot seems to have an uncanny understanding of the problem ... Other times, Copilot looks completely clueless'' \cite{DBLP:conf/sigcse/Wermelinger23}, and Jesse {\em et al.} discuss Codex's tendency to generate buggy code in some situations \cite{DBLP:conf/msr/JesseADM23}.
None of these works consider adversarial perturbation of prompts as a mechanism for hindering LLM-assisted cheating. 
%%\noindent\textbf{LLMs in Education.}
%%More broadly, Milano {\em et al.} considers three strategies for mitigating LLMs' potential negative impacts on education: (1) banning LLMs, (2) reverting to traditional methods, and (3) fully embracing them \cite{DBLP:journals/natmi/MilanoML23}, but does not consider adversarial approaches aimed at making LLMs less useful for academic dishonesty. 
Sadasivan {\em et al.}{} gives empirical evidence highlighting concerns that LLM-generated texts can easily evade current AI detection mechanisms \cite{sadasivan2023aigenerated},
underscoring the need for more advanced detection technologies that can follow the continuous advancements in LLM capabilities and ensuring the integrity of academic work. 
%%\SR{Current headings of the following paragraphs are too broad.}
%%\RY{Resolved}
% Swapping 7.3 and 7.4 for better flow of the idea

%% Sazz, commented out this for space constraints.
% \noindent
% \textbf{Understanding Decision Making of LLMs.} Understanding the nature of the black-box decision-making behaviors is crucial to attack the performance of LLMs. \cite{DBLP:conf/ccs/DangHC17} shows that morphing inputs can effectively mislead black-box models, a technique we adapted in our methodology. Additionally, the reliability and limitations of explanation techniques for black-box models such as LIME and SHAP, explored in \cite{DBLP:conf/aies/SlackHJSL20}, provide valuable insights into the transparency and interpretability of LLMs with adversarial attacks. 

\noindent
\textbf{Adversarial Attacks on Code Generation LLMs.}
Real-world applications relying on LLMs can be susceptible to vulnerabilities arising from adversarial attacks~\cite{shayegani2023survey}. 
Various strategies have been proposed to enhance the adversarial robustness of LLMs~\cite{jiang2020smart,ShettySF18,wang2021infobert}, but these methods differ significantly, and there is a lack of standardization in the adversary setups used for valuation~\cite{DBLP:conf/nips/WangXWG0GA021}.  
% Wang {\em et al.} argue that reliable benchmarks should be unambiguously annotated by humans as automatically generated examples can also fool humans \cite{DBLP:conf/emnlp/MorrisLLJQ20}, which in our case is our goal. 
%%\RY{Our goal is to fool both the model and students} 
%%To challenge these disparity, they developed a robust dataset, namely AdvGLUE. }
% Using a robust dataset developed for this purpose, Wang {\em et al.} \cite{wang2023on} explored the zero-shot robustness \cite{mao2023understanding} of ChatGPT. 
Wang {\em et al.}'s experiments show that, despite its relative dominance over other LLMs, ChatGPT's performance is nevertheless sensitive to adversarial prompts and is far from perfect when attacked by adversarial examples. To the best of our knowledge, our work is the first attempt at studying the \textit{\textbf{Robustness in Education}} with adversarial attacks. 
%%Compare to the works by \cite{DBLP:conf/nips/WangXWG0GA021} or \cite{zhu2023promptbench} that focus on generating attacks aimed to mislead specific tasks, works like \cite{zou2023universal} or \cite{chen2023LLMAttack} 
Other research showed that adversarial attacks are also effective in breaking guards against generating malicious or unethical content \cite{zou2023universal,chen2023LLMAttack}.
Incorporating the methods suggested by \cite{wang2023generating} for generating natural adversarial examples could be explored in the future.

\section{Conclusion}
High-performant LLMs pose a significant threat to enable cheating on introductory programming assignments. It investigates the potential of adversarial perturbation techniques to impede LLM-assisted cheating by designing several such methods and evaluating their efficacy in a user study. The result suggests that the combination of the perturbation indeed caused a 77\% reduction in the correctness of the generated solutions, showing early promises.
Our perturbations show positive results, but they might only be effective temporarily. Future techniques, including rigorous training data and protective layers in the prompting pipeline of LLMs, could counter these results. We hope our study will inspire ongoing efforts to prevent the misuse of LLMs in academic settings.

\section{Limitations}
\label{limitations}

\noindent
\textbf{Impact of running the user study with students exposed to the assignments.} One possible limitation of our user study is that it was conducted on students who already took CS1 and CS2 courses; thus, the finding might not hold for target students. However, as the study aimed to see if students can detect and reverse our perturbations, we hypothesize that experienced students will be more equipped to do so than new ones. Thus, if our results suggest that a given perturbation technique is effective in impeding \textit{reversal} for the study group, it is likely to be effective on the new students (actual target group) as well. However, if our results suggest that a perturbation technique is ineffective for the study group, it does not imply that it will be ineffective for the new students. This means our study offers a conservative estimation of the efficacy of the perturbation techniques on the students. Given that designing an ethically acceptable user study with new students is challenging, we argue this is acceptable. For example, Shalvi \textit{et al.}~\cite{shalvi2011justified} hypothesized that reducing people's ability to observe desired counterfactuals reduces lying. Thus, one can argue that exposing new students to the ``ChatGPT way'' of solving problems is ethically more questionable than exposing more mature students. This is because \textit{a)} The fact that they will know they can get away might incentivize cheating, as they are likely unaware of the long-term consequences. The damage is arguably less for the students with some CS fundamental knowledge and more insights into the long-term consequences.

We also want to note that even if we ignore the ethical challenge mentioned above, designing a reasonable study with new students is challenging. For example, all CS students are required to take the courses from which we took the problems, and the problems typically address concepts that have been discussed in class. So, if we wanted students who have not seen those (or similar) problems, we would have to take non-CS students who have not taken those classes and who would not have the background to solve those problems. This implies either running the study as part of the course offering or emulating the course for the study. Given the duration and volume it needs, it will be challenging to design such a study while keeping all the other confounding factors (i.e., controlling the models used) in check. Given these challenges, we chose to use the ChatGPT interface for the user study instead of an API-based tool with the trade-off between user comfort and controllability of model parameters or versions. However, seeing how the findings hold under different user settings will be interesting. Considering the complexities and numerous factors in designing such studies, they warrant dedicated independent research efforts.

\noindent
\textbf{Impact of perturbation on understandability.} Perturbations can affect understandability. Our work is intended to provide instructors with additional tools and techniques to deter LLM-assisted cheating; it is up to the instructor to ensure that any applied perturbations do not impact the clarity of the problem description.  For example, a judicious application of the ``sentence removal'' perturbation technique we describe can be combined with using images to replace the semantic content of the removed sentences.
Additionally, some perturbation techniques, such as ``unicode replacement'' and ``character removal'' may be easily reversed by a student who notices them, as our user study revealed.  Thus for these ``smart tweak'' perturbations, the key requirement is to be as imperceptible as possible, to avoid detection.
We also note that this is the first work to proactively deter the use of LLM-assisted cheating in the academic context, which is an urgent problem. It would be interesting to see what other approaches can be more effective for this purpose in the future or to run studies to find perturbations that do not affect students trying to solve problems honestly but do affect students who submit ChatGPT solutions. 
Additionally, prompts engineering to reverse the perturbation to understand their strengths can be a great complement to evaluating the strength of perturbations, together with user studies, or in cases where user studies might be infeasible to run. It would also be interesting to run follow-up studies on what factors affect comprehensibility to develop principles for designing ``understandability-preserving perturbations."
Investigating all these interesting questions can be both motivated and enabled by the current work.

\noindent
\textbf{Other limitations.}
We use \CODERL{} as the surrogate model, which might not be a close approximation of the target models. Despite this limitation, \CODERL{} is successful in generating perturbed samples to run our field study.  Finally, we ran the user study with only 6 assignments, which might hurt the generalizability of the findings. ChatGPT provides personalized answers, which might cause variances in our results. To counter this, we added redundancy in our study design and reported average results. 

\section{Ethical Considerations} Our study was approved by the IRB of the designated institute. We recruited students who have already taken CS1 and CS2 to avoid academic integrity violations. Participants were compensated with a reward of \$20 for their contribution. During the user study, we did not collect any personally identifiable data. Lastly, all the experiments on \GPT{} and \Mistral{} models were done with premium API access. We also used GitHub Copilot under an academic subscription to ensure fair and responsible use. The replication package, which includes the data and source code, will be available to researchers on request.

\section*{Acknowledgements}

We thank Genesis Elizabeth Benedith and Logan Michael Sandlin for their involvement during the initial stage of the project. We also thank the instructors of CS1 and CS2 who taught these courses at the University of Arizona over the years, including Adriana Picoral, 
Janalee O'Bagy, Reyan Ahmed, Russell Lewis, Todd Proebsting, and Xinchen Yu, for sharing the assignments, solutions, and syllabus with us. Finally, we thank the anonymous reviewers for their feedback on the initial draft of the paper.

\bibliography{bibliography}

\appendix

\section{Syllabus of CS1}\label{appen:syllabus_cs1}
% https://xinchenyu.github.io/csc110-spring2024/syllabus.html

% \subsection{Instructor}
% Xinchen Yu, Assistant Professor of Practice

\subsection{Course Description}
An introduction to programming with an emphasis on solving problems drawn from a variety of domains. Topics include basic control and data structures, problem-solving strategies, and software development tools and techniques. Specifically, the Python programming language will be taught.

\subsection{Course Objectives}
By the end of the semester, you should be able to write complete, well-structured programs in Python.

\subsection{Expected Learning Outcomes}
Students who successfully complete this course should be able to:

\begin{itemize}
	\item Use variables, control structures, basic data types, lists, dictionaries, file I/O, and functions to write correct 100 - 200 line programs.
	\item Decompose a problem into an appropriate set of functions, loops, conditionals, and/or other control flow.
	\item Find bugs when code is not working as expected using print statements and computational thinking skills, and will be able to understand and resolve errors.
	\item Write clean, well-structured, and readable code.
	\item Follow a provided style guide to write clean, well-structured, and readable code.
	\item Explain the conceptual memory model underlying the data types covered in class and demonstrate the ability to convert integers and text to and from binary.
\end{itemize}

\section{Syllabus of CS2}\label{appen:syllabus_cs2}
% https://obagy.com/cs120/csc120-fall24syllabus.pdf
% https://www2.cs.arizona.edu/~abureyanahmed/Syllabus_csc120_fall2024.pdf

% \subsection{Instructor}
% Reyan Ahmed, Assistant Professor of Practice

\subsection{Course Description}
This course provides a continuing introduction to programming with an emphasis on problem-solving. It considers problems drawn from various domains (including Computer Science). It emphasizes both the broader applicability of the relevant data structures and programming concepts, as well as the implementation of those structures and concepts in software. Topics include arrays, lists, stacks, queues, trees, searching and sorting, exceptions, classes and objects; asymptotic complexity; testing, and debugging. 

% This course provides a continuing introduction to programming with an emphasis on problem-solving. It considers problems drawn from a variety of domains, including Computer Science, and emphasizes both the broader applicability of the relevant data structures and programming concepts, as well as the implementation of those structures and concepts in software. Topics include: arrays, lists, stacks, queues, trees, searching and sorting, and exceptions; classes and objects; invariants and pre-/post-conditions; incremental program development, testing, and debugging.

% \textbf{Note:} This class has a significant programming component.

\subsection{Course Objectives}
The course will provide a foundation in fundamental computer science concepts such as object-oriented programming, data structures and abstract data types, asymptotic worst-case complexity, program design, testing, and debugging. 

% This course focuses on programming in the larger context of problem-solving, with problems drawn from a variety of domains including Computer Science itself. Like its predecessor, this course is taught in Python, thereby allowing students to get more comfortable with programming and with using programs to solve problems. Additionally, it explores the broader applications of programming concepts, including data structures such as lists, stacks, queues, etc., and discusses the software implementation of those concepts.

\subsection{Expected Learning Outcomes}
Students who successfully complete this course should be able to:
\begin{itemize}
    \item Effectively decompose simple programming problems into suitable functions.
    \item Comfortably write moderate-sized (100–300 line) programs incorporating a variety of control and data structures.
    \item Implement common data structures such as stacks, queues, linked lists, and trees and use recursive solutions when appropriate;
    \item Implement classes given design guidance;
    \item Use a provided style guide to produce clean, readable code;
    \item Identify and create black box and white box tests and use assertions to facilitate the testing and debugging of their programs;
    \item Determine the time complexity of simple algorithms and state their complexity in terms of big-O notation.
\end{itemize}
% \begin{itemize}
% 	\item Effectively decompose simple programming problems.
% 	\item Comfortably write moderate-sized (100–300 line) programs incorporating a variety of control and data structures.
% 	\item Debug and test programs.
% \end{itemize}

\section{Short and Long Problems}\label{appen:short:long}

Figure \ref{fig:short_and_long_problem} shows an example of short and long problems.

\begin{figure}[!ht]
\centering
\begin{lstlisting}
In a file jaccard.py write a function jaccard(set1, set2) that takes as arguments two sets set1 and set2 and returns a floating-point value that is the Jaccard similarity index between set1 and set2. The definition of the Jaccard similarity index is (see also: Section 2.B of the long problem spec; Wikipedia):
similarity(set1, set2)  = | set1 @$\cap$@ set2 | / | set1 @$\cup$@ set2 |

If set1 and set2 are both empty sets, their similarity is defined to be 1.0.

Examples
set1 set2 jaccard(set1, set2)
{'aaa', 'bbb', 'ccc', 'ddd'} {'aaa', 'ccc'} 0.5
{1, 2, 3} {2, 3, 4, 5} 0.4
{1, 2, 3} {4, 5, 6} 0.0
\end{lstlisting}
% \vspace{-10pt}
$(a)$ Short problem

\begin{lstlisting}
In a file update_board.py write the following functions:
update_board(board, mov): board is an internal representation of a board position, mov is a tuple of integers specifying a move. It returns the internal representation of the board resulting from making the move mov in board board.

update_board_interface(board_str, mov): board_str is an external representation of a board position (a string of 0s and 1s), mov is a tuple of integers specifying a move. This function converts board_str to your internal representation of a board position, calls your function update_board() described above, converts the value returned by update_board() to an external representation of a board (a string of 0s and 1s), and returns the resulting string. This function thus serves as the external interface to your update_board() function.

2.3.2. Examples
board_str mov update_board_interface(board_str, mov)
110001100101011 (14, 13, 12) 110001100101100
110001100101011 (0, 1, 3) 000101100101011
0110011011 (5, 2, 0) 1100001011
\end{lstlisting}
% \vspace{-10pt}
$(b)$ Long problem

\caption{Examples of short and long problems}
\label{fig:short_and_long_problem}
\end{figure}

\begin{figure}[!ht]
    \centering
    \includegraphics[width=0.7\linewidth]{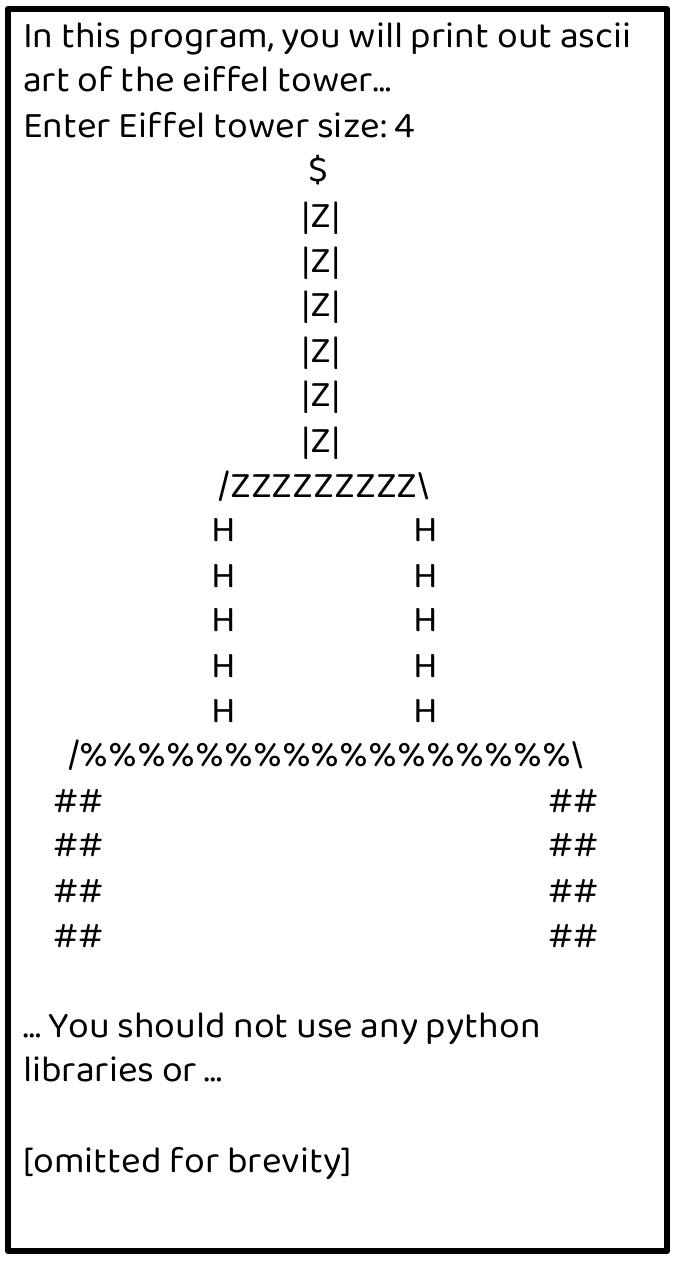}
    % \vspace{-20pt}
    \caption{An example CS1 problem where \CODERL, \GPT and \GITHUBCOPILOT~scored 0\%.}
    \label{fig:cs1:problem}
\end{figure}

\section{LLM Code Generation Methodology} \label{app:methodology}
\paragraph{\CODERL.}
% Generating code with \CODERL~ proves to be a more straightforward process compared to \GITHUBCOPILOT, mainly due to the availability of APIs that eliminate the need for GUI automation, such as VS Code. 
% Generating code with \CODERL~ is easier than using \GITHUBCOPILOT~ because \CODERL~ has APIs that eliminate the need for GUI automation like VS Code.
To initiate code generation with \CODERL, we first create an instance of the tokenizer and model using the HuggingFace API. To ensure obtaining the best solution, we set the \textit{temperature} to $0$ and the output token limit to its maximum allowable limit. Then, we tokenize the prompt and send it to the model. The model generates a list of tokens from the given prompt of tokens. After detokenizing the output, we get a source code, which serves as the solution to the given assignment problem.% This streamlined approach, facilitated by accessible APIs, makes code generation with \CODERL~ efficient and effective.

\paragraph{\GITHUBCOPILOT.}
To generate code with \COPILOT, we employ PyAutoGUI to automate VS Code. The step-by-step process starts with opening VS Code in a new window and creating a new Python file. We paste the prompt into the file, surrounded by a docstring comment. Next, we ask \COPILOT{} to generate multiple variations of code in a new window using the custom keyboard shortcut. Then, we close the VS Code after saving the responses in separate files.
The subsequent steps vary based on the type of problem. For short problems, we handle cases where the code can either be a standalone program generating output or a function/class definition. In the latter case, the code generation is done for that specific code. Conversely, for standalone programs, we add the 
% ``\mintinline{python}{if __name__ == '__main__':}'' --- 
``\lstinline[basicstyle=\normalsize\ttfamily, language=python, breaklines=false]{if __name__ == '__main__':}'' 
block at the bottom of the file and let \COPILOT{} call the generated function/class. At this point, \COPILOT{} provides inline suggestions rather than separate windows for alternatives.
For longer problems, we reopen the generated code in VS Code and allow \COPILOT{} to provide up to 15 inline suggestions. However, if \COPILOT{} generates its own 
% ``\mintinline{python}{if __name__ == '__main__':}'' --- 
``\lstinline[basicstyle=\normalsize\ttfamily, language=python, breaklines=false]{if __name__ == '__main__':}'' 
block, we stop, as further code generation may lead to uncompilable results.

As both short and long problems can generate up to 10 solutions for a single prompt, we run all generated solutions through autograders and select the one with the highest score for evaluation. This methodology ensures efficient code generation and selection of the most appropriate solution for the given prompt.

% \begin{figure}[!ht]
% \centering
% \begin{minted}[frame=single, breaklines=true, breakanywhere=true, fontsize=\scriptsize]{text}
% Write a Python program that does the following:

% <problem statement>

% Please omit any explanations of the code.
% \end{minted}
% \vspace{-10pt}
% \caption{Prompt to generate source code from \GPT~}
% \label{fig:prompt}
% \end{figure}

\begin{figure}[!ht]
\centering
\begin{lstlisting}
Write a Python program that does the following:

<problem statement>

Please omit any explanations of the code.
\end{lstlisting}
% \vspace{-10pt}
\caption{Prompt to generate source code from \GPT~}
\label{fig:prompt}
\end{figure}

\paragraph{\GPT.}
We use the OpenAI API to generate code using \GPT. Specifically, we use the \texttt{gpt-3.5-turbo-0301} model to ensure consistency throughout our experiments. Similar to \CODERL, we set the \textit{temperature} to 0 to obtain the most optimal source code deterministically. Since \GPT~ is a general-purpose language model not specifically designed for code generation only, we add qualifying sentences around the prompt instructing \GPT{} to omit explanations and produce only code (since non-code explanatory text could induce syntax errors in the autograder). Figure \ref{fig:prompt} shows the prompt we use to generate code from \GPT. This way, we exclusively receive code outputs from the model.

\paragraph{\Mistral.}
We used the Mistral API to generate code using \Mistral. Specifically, we used the \texttt{mistral-large-2402} model to ensure consistency throughout our experiments. Because Mistral's API is very similar to OpenAI's API, we followed the same methodology and used the same model parameters to interact with the API.

\paragraph{\CLLaMA.}
We used Ollama, a lightweight and extensible framework for running LLMs on local machines, to host the CodeLlama-7b-instruct model based on Meta's Llama 2. The instruct model was chosen as it is trained to output human-like answers to given queries, which we believed to be closest to ChatGPT in terms of the generated solutions. The steps include installing Ollama and simply calling \textbf{ollama run codellama:7b-instruct `<prompt>'} to generate the outputs. To the best of our knowledge, there isn't a straightforward way to tweak the parameters of the models from the provided user manuals, so we used the default model. Although the generated answers often contained comment blocks as well as codes, most outputs wrapped the code blocks with identifiable texts such as \textbf{'''}, \textbf{[PYTHON]} or \textbf{```python}, we extracted the codes accordingly. Otherwise, we simply used the generated output.

\section{Descripiton of our Perturbation Techniques}\label{app:perturb}

\subsection{Core perturbations.}
 
\noindent
\textbf{Token (remove):} Breaking subword tokens profoundly impacts LLM performance~\cite{DBLP:conf/emnlp/LiuYHLLMYW22, DBLP:conf/nips/WangXWG0GA021}. By consulting SHAP, in this technique, we remove the top 5 tokens 
%\sout{separately from the assignment description and create 5 independent perturbed variants} \textcolor{red}{
from the assignment description and create 1 perturbed variant of a given assignment. We generated 63 short and 12 long variants in total.

\noindent
\textbf{Character (remove):} Following the same principle as \textit{Token (remove)} to break subwords, in this perturbation technique, we remove a random character from each of the top 5 tokens to create 1 variant. We generated 63 short and 12 long variants in total.

\noindent
\textbf{Random (insert):} To break subwords, we also design another perturbation by inserting redundant characters, such as hyphens and underscores, in the top 5 tokens; similarly, we generate 1 variant of inserting redundant characters, such as hyphens and underscores, into the top tokens in the assignments. We generated 63 short and 12 long variants in total.

\noindent
\textbf{Sentence (remove):} For sentence removal, we remove a third of the sentence from the assignment description sequentially. We chose one-third so as to not remove too much relevant information, and we removed sequential sentences to create a large hole in the information provided to the models. If the assignment description has less than 3 sentences, we remove only 1 sentence. This produces a variable number of perturbed variants. We generated 594 short and 857 long variants in total.
%\SR{@Alex, it will be great if you mention, how many variants we have in total for all the assignments.}
%\AC{Done}
%\AC{We have 64 short problems for CS2, but SHAP did not work for one of the problems, and I was never able to fix it, so for RQ2 we only morphed 63 of the short problems. So in the above paragraphs when I mention how many variants for each morphing technique, the techniques that generate just one variant only have 63 short problem variants.}

\noindent
\textbf{Sentence (rephrase):} Rephrasing of sentences is known to be effective in degrading LLM performance~\cite{DBLP:conf/ijcnlp/XuCBV22, DBLP:conf/emnlp/MorrisLLJQ20, DBLP:conf/emnlp/AlzantotSEHSC18, DBLP:conf/nips/WangXWG0GA021}. Thus, we leverage rephrasing sentences to design this perturbation. %\sout{Similar to \textit{Sentence (remove)} perturbation, here, too, we rephrase the top 3 sentences separately to create 3 independent variants.} 
First, we rank the sentences by accumulating the Shapley values of the tokens corresponding to a given sentence; then, we remove the top 3 sentences to create 3 independent variants. We use \GPT to obtain high-quality phrases. We generated 177 short and 32 long variants in total.

\noindent
\textbf{Token (synonym):} Tokens are the building blocks of language models, which have been used as perturbation units in context~\cite{Boucher023, DBLP:conf/dasc/Al-EssaAAM22, DBLP:conf/nips/WangXWG0GA021}. Therefore, we design a perturbation technique. to substitute tokens with their synonyms. Specifically, we replace the top 5 tokens from the SHAP with their synonyms to create 5 different variants. For each top-ranked token, we replace all instances of that token in the prompt with its synonym, even if other occurrences are not top-ranked. We do this to ensure that if the token provides necessary information to the model, it cannot be obtained from another token occurrence in the assignment description.  We generate contextual synonyms for a given token using \GPT. We provide the sentence containing the token as the context for the \GPT~ model and ask for synonyms for the token. We generated 1836 short and 216 long variants in total.

\begin{figure}[ht]
    \centering
    \includegraphics[width=\linewidth]{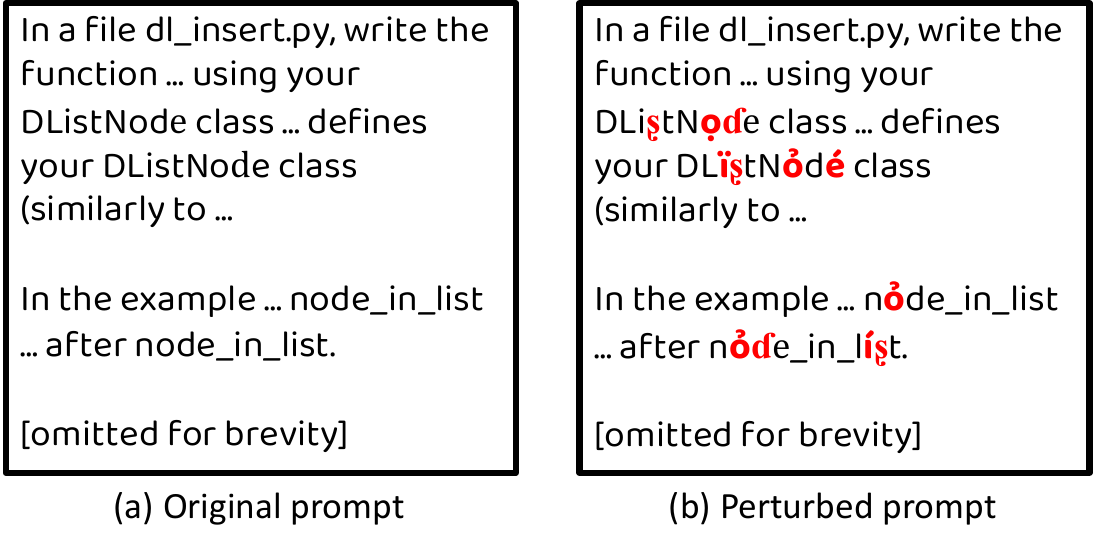}
    % \vspace{-20pt}
    \caption{Replacing 12 characters for 5 tokens with their Unicode lookalike from an assignment prompt caused correctness scores to drop from 100\% to 0\% in \GPT.}
    \label{fig:intro_motivation_figure_2}
\end{figure}

\noindent
\textbf{Token (unicode):} Recent research shows that adversarial attacks can be effective even in a black-box setting without visually altering the inputs in ways noticeable to humans, which includes replacing characters with Unicode lookalikes \cite{ShettySF18, BoucherS0P22}. To leverage this, we create a perturbation method to replace characters in the top 5 tokens (from SHAP) with their Unicode lookalikes to create 1 variant (Figure~\ref{fig:intro_motivation_figure_2}). We generated 63 short and 12 long variants in total.

\subsection{Exploratory Perturbations.}

\noindent
\textbf{Tokens (synonym):} To understand the potential of synonym-based perturbation, we create a new type of perturbation method to replace the top 5 tokens from the SHAP with their synonyms to create 5 different variants. However, we do not replace the top-ranked occurrences of a given token -- not all occurrences in a given assignment prompt. We generated 2373 short and 223 long variants in total.

\noindent
\textbf{Prompt (Unicode):} Similarly, to study the full potential of substituting characters with Unicode lookalikes, we apply it to the whole assignment statement under this technique. We recognize that this perturbation might easily get noticed; however, we add it to understand how detectability might impact the actual performance in the field study. We generated 63 short and 12 long variants in total.

\noindent
\textbf{Random (replace):} Existing studies show evidence that LLMs are prone to memorizing training data~\cite{DBLP:journals/corr/abs-2112-12938, DBLP:conf/uss/CarliniTWJHLRBS21, DBLP:conf/iclr/CarliniIJLTZ23}. Thus, these models are highly sensitive to input variations, and even slight changes in the prompt may lead to substantial differences in the generated output~\cite{DBLP:conf/iclr/ZhangLCDBTHC22, DBLP:conf/acl/Jin0SC022, DBLP:conf/chi/ReynoldsM21}. Under this hypothesis, replacing specific tokens with random strings may significantly influence performance, as such substitution may alter the context~\cite{DBLP:conf/icml/ShiCMSDCSZ23, DBLP:journals/corr/abs-2307-03172, DBLP:conf/nips/WangXWG0GA021}. We design a new exploratory perturbation technique to leverage this insight. Under this technique, we tweak assignments by replacing file names, function names, and class names specified in the problem statement with random strings, where these names are discovered manually. We store the original names and random strings, then in the code generated by the models, replace the instances of the random strings with the original names. This is to make sure that the autograders don't give a score of 0 for a good solution that uses the random string. We generated 63 short and 12 long variants in total.

% \section{Perturbation}

% \begin{table}[h]
%     \caption{Average complexity of solutions where morphing failed and succeeded}
%     \label{tab:morph_failed}
%     \resizebox{\columnwidth}{!}{%
%         \begin{tabular}{|c|c|c|c|}
%             \hline
%             Category & Number of Lines & Token Count & Cyclommatic Complexity \\
%             \hline
%             Failed & 7.6 & 67.3 & 3.0 \\
%             \hline
%             Succeeded & 14.3 & 105.0 & 2.0 \\
%             \hline
%         \end{tabular}
%     }
% \end{table}

\section{User Study}\label{app:userstudy}

% Please add the following required packages to your document preamble:
% \usepackage{graphicx}
\begin{table}[!ht]
\caption{Demography of the participants}
\label{tab:participant_demography}
\resizebox{\columnwidth}{!}{%
\begin{tabular}{|c|c|c|c|}
\hline
\textbf{Participants} &
  \textbf{\begin{tabular}[c]{@{}c@{}}Academic\\ Status\end{tabular}} &
  \textbf{\begin{tabular}[c]{@{}c@{}}Proficiency in Python\\ (out of 5)\end{tabular}} &
  \textbf{\begin{tabular}[c]{@{}c@{}}LLM Usage Frequency\\ (weekly)\end{tabular}} \\ \hline
P1 & Junior    & 5 & Occasionally (3-5 times)             \\ \hline
P2 & Junior    & 4 & Never                                \\ \hline
P3 & Senior    & 5 & Occasionally (3-5 times)             \\ \hline
P4 & Senior    & 5 & Occasionally (3-5 times)             \\ \hline
P5 & Senior    & 5 & Very frequently (More than 10 times) \\ \hline
P6 & Senior    & 4 & Rarely (1-2 times)                   \\ \hline
P7 & Sophomore & 4 & Occasionally (3-5 times)             \\ \hline
P8 & Senior    & 4 & Very frequently (More than 10 times) \\ \hline
P9 & Sophomore & 4 & Occasionally (3-5 times)             \\ \hline
P10 & Senior    & 4 & Occasionally (3-5 times)             \\ \hline
P11 & Senior    & 4 & Regularly (6-10 times)               \\ \hline
P12 & Senior    & 4 & Rarely (1-2 times)                   \\ \hline
P13 & Sophomore & 5 & Occasionally (3-5 times)             \\ \hline
P14 & Senior    & 4 & Rarely (1-2 times)                   \\ \hline
P15 & Junior    & 4 & Rarely (1-2 times)                   \\ \hline
P16 & Senior    & 4 & Rarely (1-2 times)                   \\ \hline
P17 & Junior    & 4 & Occasionally (3-5 times)             \\ \hline
P18 & Junior    & 4 & Occasionally (3-5 times)             \\ \hline
P19 & Sophomore & 4 & Never                                \\ \hline
P20 & Junior    & 3 & Never                                \\ \hline
P21 & Junior    & 5 & Rarely (1-2 times)                   \\ \hline
P22 & Senior    & 4 & Never                                \\ \hline
P23 & Junior    & 3 & Rarely (1-2 times)                   \\ \hline
P24 & Senior    & 5 & Very frequently (More than 10 times) \\ \hline
P25 & Senior    & 4 & Never                                \\ \hline
P26 & Senior    & 4 & Regularly (6-10 times)               \\ \hline
P27 & Junior    & 4 & Occasionally (3-5 times)             \\ \hline
P28 & Junior    & 3 & Rarely (1-2 times)                   \\ \hline
P29 & Senior    & 4 & Very frequently (More than 10 times) \\ \hline
P30 & Senior    & 4 & Regularly (6-10 times)               \\ \hline
\end{tabular}%
}
\end{table}

\begin{table}[!ht]
\centering
\caption{User Study Questions}\label{tab:userstudy:questions}
\resizebox{\columnwidth}{!}{%
\begin{tabular}{|p{1.5\columnwidth}|}
\hline
\textbf{Questions} \\ \hline
How proficient are you in the Python programming language? \\ \hline
How hard did the problem seem to you while you were solving it? (For each problem) \\ \hline
How much time (in minutes) did you spend on this problem? (For each problem) \\ \hline
How did you validate the ChatGPT-generated solutions? (For each problem) \\ \hline
Did you notice anything unusual about the problem statement? (For each problem) \\ \hline
How did you avoid the ``unusualness'' in the problem statement while solving the problem? (For each problem) \\ \hline
On average, how many hours do you dedicate to coding or problem-solving per week? \\ \hline
How often do you utilize ChatGPT or any other Large Language Model to solve problems on a weekly basis, on average? \\ \hline
What other Large Language Models do you use or previously used? \\ \hline
\end{tabular}
}
\label{table:user_study_questions}
\end{table}

\begin{table}[!ht]
\caption{Distributions of the perturbation techniques and the problems in the user study}
\label{tab:prob_morph_dist}
\resizebox{\columnwidth}{!}{%
\begin{tabular}{l|clcc}
\cline{1-2}
\textbf{Perturbations}                  & \textbf{\#Participants} &                       &                         &                         \\ \cline{1-2} \cline{4-5} 
\texttt{Prompt (original)} & 18 & \multicolumn{1}{c}{} & \multicolumn{1}{c|}{\textbf{Problems}} & \multicolumn{1}{c}{\textbf{\# Participants}} \\ \cline{1-2} \cline{4-5} 
\texttt{Character (remove)} & 12 & \multicolumn{1}{c}{} & \multicolumn{1}{c|}{p1} & \multicolumn{1}{c}{22} \\ \cline{1-2} \cline{4-5} 
\texttt{Token (unicode)} & 13 & \multicolumn{1}{c}{} & \multicolumn{1}{c|}{p2} & \multicolumn{1}{c}{17} \\ \cline{1-2} \cline{4-5} 
% \texttt{Prompt (unicode)} & 16    & \multicolumn{1}{c}{} & \multicolumn{1}{c|}{p1} & \multicolumn{1}{c}{22} \\ \cline{1-2} \cline{4-5} 
\texttt{Tokens (remove)}      & 7     & \multicolumn{1}{c}{} & \multicolumn{1}{c|}{p3} & \multicolumn{1}{c}{13} \\ \cline{1-2} \cline{4-5} 
% \texttt{Random (replace)}     & 11    & \multicolumn{1}{c}{} & \multicolumn{1}{c|}{p3} & \multicolumn{1}{c}{13} \\ \cline{1-2} \cline{4-5} 
\texttt{Sentences (rephrase)}   & 3    & \multicolumn{1}{c}{} & \multicolumn{1}{c|}{p4} & \multicolumn{1}{c}{13} \\ \cline{1-2} \cline{4-5} 
\texttt{Sentences (remove)} & 10     & \multicolumn{1}{c}{} & \multicolumn{1}{c|}{p5} & \multicolumn{1}{c}{13} \\ \cline{1-2} \cline{4-5} 
\texttt{Prompt (unicode)}   & 16    & \multicolumn{1}{c}{} & \multicolumn{1}{c|}{p6} & \multicolumn{1}{c}{12} \\ \cline{1-2} \cline{4-5} 
\texttt{Random (replace)}                  & 11    &                       &                         &                         \\ \cline{1-2}
\end{tabular}%
}
\end{table}

\subsection{Description of the thematic analysis}
% \SR{Where to add the following text? Consensus-based resolution is considered important in qualitative studies to produce meaningful insights. In our case, there were 4 disagreements between the two raters while labeling all 30 participant’s data. Which indicates that the subjective bias was minimal. We will add more insights into our thematic analysis.}
This approach consists of multiple stages. First, we familiarize ourselves with the collected data. We manually go through 50\% (15 out of 30) responses in this stage. This allows us to perform inductive coding to identify potential codes for further analysis. In the second stage, two authors generated 16 initial codes based on their familiarity with the data. These codes are data-driven and help organize information into meaningful units. Two authors assign codes to the participants' responses to the specific questions. This coding stage is done manually. To address disagreements, the authors facilitated a consensus-based resolution while combining their coding assignments. Consensus-based resolution is considered important in qualitative studies to produce meaningful insights. In our case, there were 4 disagreements between the two raters while labeling all 30 participant’s data. After that, one of the authors reviews the students' responses and corresponding conversations with ChatGPT to get the most information and update the coding. This step is iterative until saturation. We consider the coding to be saturated if no new code is assigned to the responses. Lastly, the other author validates the final coding to avoid potential bias. In the third stage, after coding the data, we start searching for themes by bringing together material under the same codes. This involves considering how codes may form broader themes that are organized hierarchically. In the fourth stage, we review and refine the potential themes. 

\begin{figure}[!ht]
    \centering
    \includegraphics[width=\linewidth]{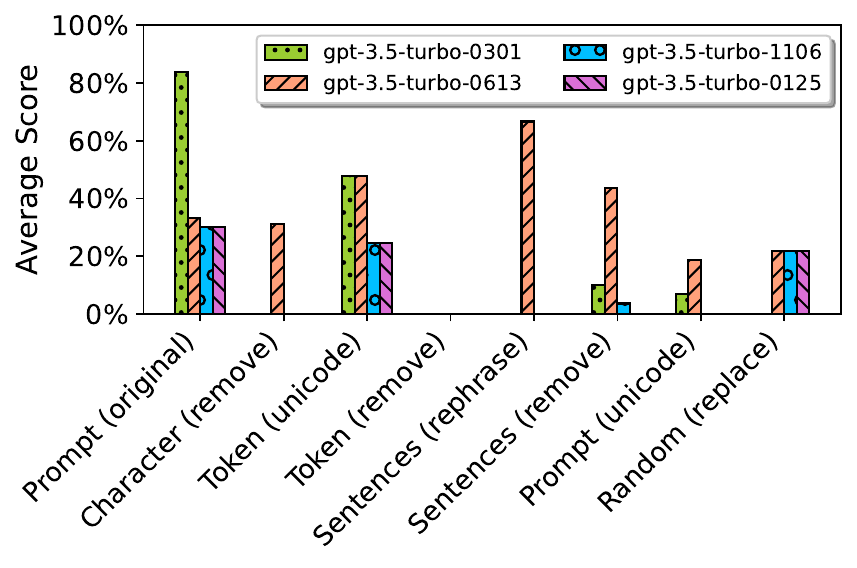}
    \vspace{-25pt}
    \caption{Average correctness score of the ChatGPT model checkpoints on the user study problems for the perturbation techniques.}
    \label{fig:rq2_model_checkpoint_average_score_barplot}
\end{figure}

\smallskip
\noindent
\textbf{Codebook for neutralizing perturbations:}
\begin{footnotesize}
\begin{itemize}
    \item Update the given problem statement
    \item Rely on ChatGPT to avoid any perturbation
    \item Did not notice anything ``unusualness''
    \item Rewrite the whole solution manually as the ChatGPT-generated solution is incorrect
    \item Rewrite a part of the solution manually
\end{itemize}
\end{footnotesize}
\noindent
\textbf{Themes and codes for validation:}
\begin{itemize}
    \item Inspecting the generated code
    \begin{footnotesize}
    \begin{itemize}
        \item Inspect the generated code without running
        \item Inspect the generated code by running
        \item Use given test cases
        \item Use manually created test cases
        \item Use ChatGPT-generated test cases
        \item Validate the solution using ChatGPT
        \item Compare to the manually written code
    \end{itemize}
    \end{footnotesize}
    \item Fixing the generated code
    \begin{footnotesize}
    \begin{itemize}
        \item Fix the code manually
        \item Fix the code using ChatGPT
    \end{itemize}
    \end{footnotesize}
    \item Verdict about the correctness
    \begin{footnotesize}
    \begin{itemize}
        \item Correct solution from ChatGPT
        \item Incorrect solution from ChatGPT
    \end{itemize}
    \end{footnotesize}
\end{itemize}

\begin{figure*}[!ht]
    \centering
    \includegraphics[width=\linewidth]{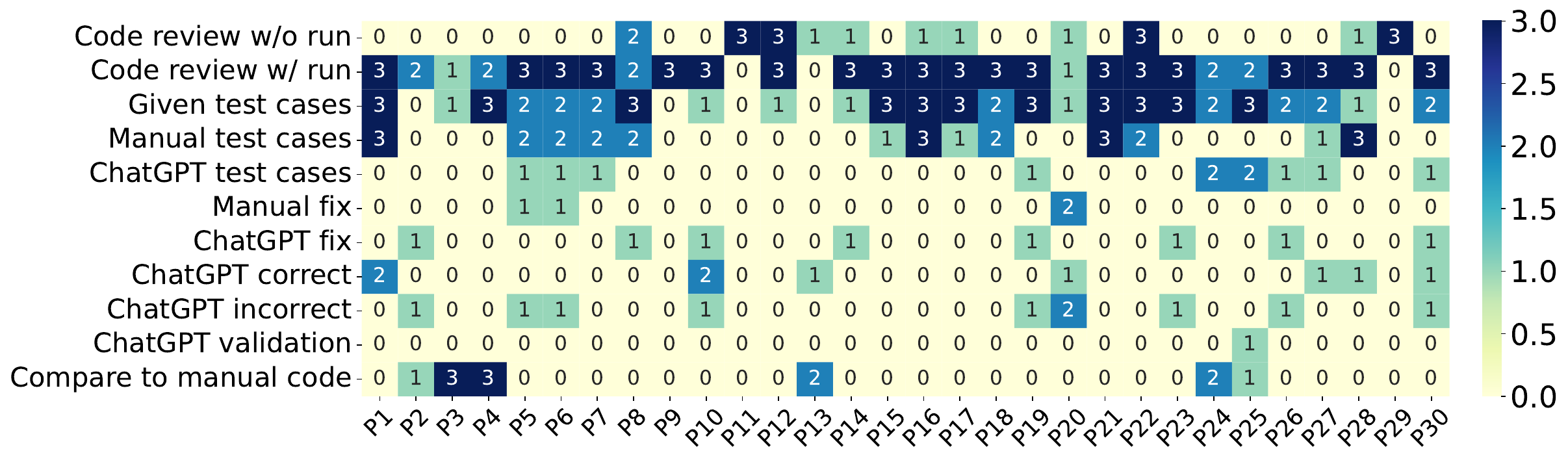}
    \caption{The vertical axis lists the most frequent validation strategies, while the horizontal axis represents participants. Each cell's value, capped at 3, indicates the number of times a specific code was applied to a participant's response across three problems. The color gradient ranges from bright yellow (indicating 0 occurrences) to dark blue (indicating 3 occurrences).}
    \label{fig:rq3_q1_coding_heatmap_per_user}
\end{figure*}

\begin{figure}[!ht]
    \centering
    \includegraphics[width=\linewidth]{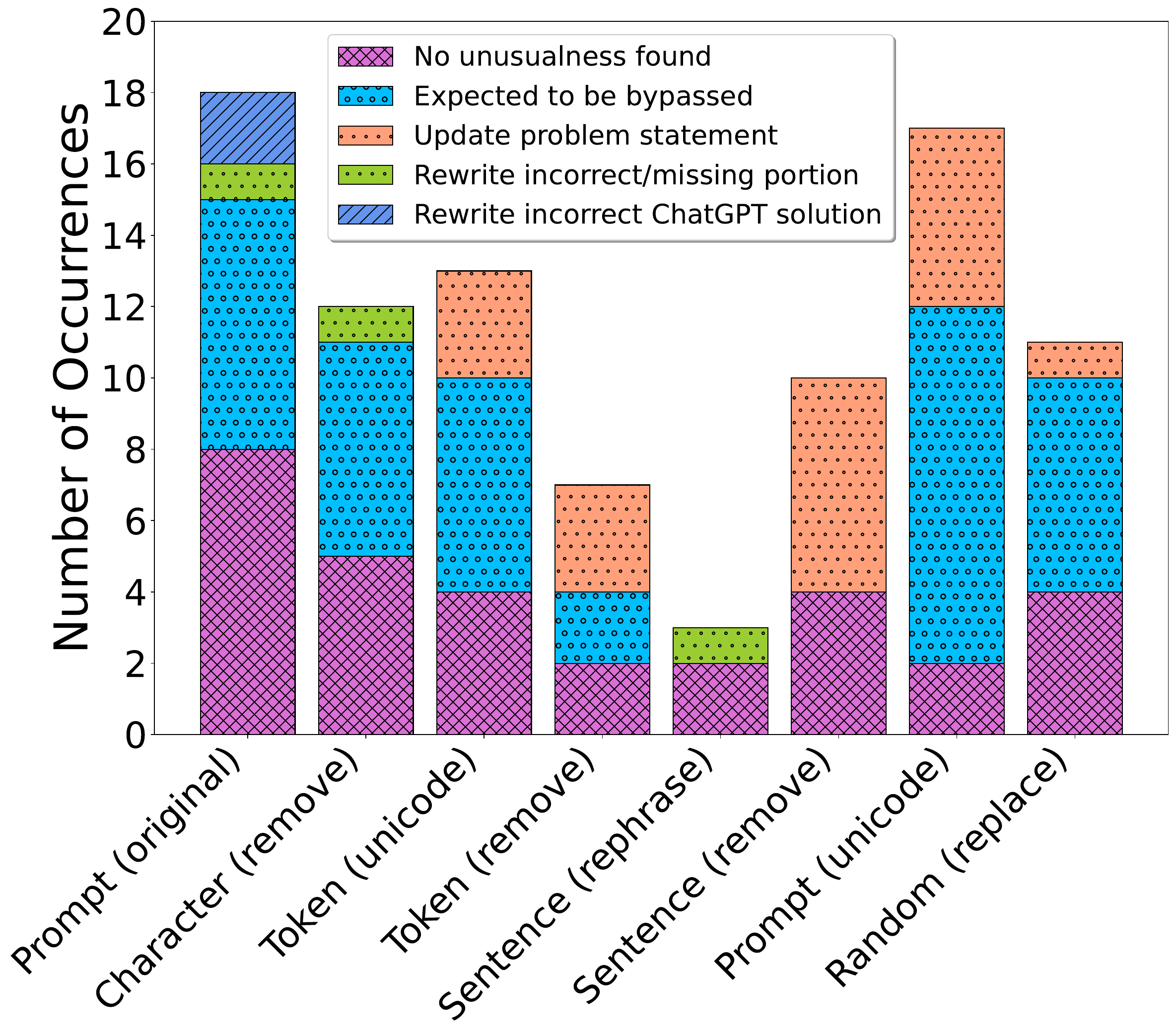}
    \vspace{-25pt}
    \caption{Number of occurrences of handling strategies for each perturbation technique.}
    \label{fig:rq3_q2_coding_vs_morphing}
\end{figure}

% \includepdf[pages=-]{user_study_form.pdf}

\section{Research Participant Agreement}\label{sec:research_participant_agreement}

\subsection{Voluntary Participation}
You are being asked to participate in a research study. Your participation in this research study is voluntary. You may choose to voluntarily discontinue participation in the study at any time without penalty, even after starting the survey. This document contains important information about this study and what to expect if you decide to participate. Please consider the information carefully. Feel free to ask questions before deciding whether to participate. 

Through this study, we will understand how well we can solve CS1 and CS2-level programming tasks using AI tools such as ChatGPT. The survey consists of three CS introductory assignment problems for each student. For each problem, you have to solve it using ChatGPT and then answer the follow-up questions. We estimate that the whole process will take around 45-60 minutes. You are free to take the survey anywhere you choose. You will be emailed the survey to complete, and you will need to provide your email address in the survey. 

By signing up you are agreeing that you took CS1 and CS2. You will proceed with the study once the verification of your historical enrollment in the CS1 and CS2 courses is confirmed with the moderator of the CS undergraduate listserv (Martin Marquez, Director of Academic and Support Services, CS). Education records used by this research project are education records as defined and protected by the Family Educational Rights and Privacy Act (FERPA). FERPA is a federal law that protects the privacy of student education records. Your consent gives the researcher permission to access the records identified above for research purposes.

\subsection{Risks for the Participants}

\begin{enumerate}
    \item \textbf{Social risk:} A minor risk is the potential of loss of confidentiality because the form asks for your email address. Google Forms automatically collects email addresses for the survey, so the email address will be attached to the survey responses.
    \item \textbf{Economic risk:} An economic risk may be that you complete the vast majority of the survey, but we cannot reward any cash, and so you lose some leisure time with no cash reward.
    \item \textbf{Psychological risk:} A psychological risk may be that you may get fatigued while solving the given problems.
\end{enumerate}

However, the risks here are largely minimal. The analysis considers the survey responses as a whole and does not investigate one specific survey response. That said, your email address will be removed before the analysis of the surveys after you collect your reward (details below).

\subsection{Incentive}

You will receive a \$20 Amazon e-gift card for completing the survey in full. To receive your \$20 award, please contact the Anonymized author. He will then check that you have completed the survey in full using your email and arrange the payment. You must collect your reward within one month of completing the survey. For any compensation you receive, we are required to obtain identifiable information such as your name and address for financial compliance purposes. However, your name will not be used in any report or analysis of the survey results. Identifiable research data will be stored on a password-secured local lab computer accessible only to the research project members.

\subsection{Confidentiality of Data}
Your information may be used for future research or shared with another researcher for future research studies without additional consent. In addition, your email addresses will be deleted from the response spreadsheets, which will be stored
on a password-secured local server computer accessible only by the research team members. The form containing the list of student emails that signed up to participate will be deleted once all surveys are complete. Once the entire research project is complete and the conference paper is published, anyone can view the results of the survey by referring to the conference website. The conference at which this paper will be accepted cannot be guaranteed at this moment.

The information that you provide in the study will be handled confidentially. However, there may be circumstances where this information must be released or shared as required by law. The Institutional Review Board may review the research records for monitoring purposes.

For questions, concerns, or complaints about the study, you may contact the Anonymized author. By completing the entire survey, you are allowing your responses to be used for research purposes.

\subsection{Instructions to the Participants}\label{sec:instructions_to_the_participants}

\begin{enumerate}
    \item Create a free ChatGPT (3.5) account if you don't have any.
    \item Each problem comes with a problem statement (shared via email). Create a separate chat window in ChatGPT to solve each problem.
    \item After solving each problem, you have to answer the corresponding survey questions.
    \item You also have to give the shareable link of the chat from ChatGPT for each problem. \href{https://help.openai.com/en/articles/7925741-chatgpt-shared-links-faq}{(ChatGPT Shared Links FAQ)}
    \item Don't delete the chats until you receive an email from us about the deletion step.
\end{enumerate}

\clearpage

\end{document}